\documentclass[letterpaper]{article} % DO NOT CHANGE THIS
\usepackage{aaai2026}  % DO NOT CHANGE THIS
\usepackage{times}  % DO NOT CHANGE THIS
\usepackage{helvet}  % DO NOT CHANGE THIS
\usepackage{courier}  % DO NOT CHANGE THIS
\usepackage[hyphens]{url}  % DO NOT CHANGE THIS
\usepackage{graphicx} % DO NOT CHANGE THIS
\usepackage{natbib}  % DO NOT CHANGE THIS AND DO NOT ADD ANY OPTIONS TO IT
\usepackage{caption} % DO NOT CHANGE THIS AND DO NOT ADD ANY OPTIONS TO IT
\usepackage{algorithmic}
\usepackage{algorithm}
\usepackage{multirow}

\usepackage{booktabs}       % professional-quality tables
\usepackage{amsfonts}       % blackboard math symbols
\usepackage{amsthm}         % theorem environments
\usepackage{amsmath, amssymb}
\usepackage{nicefrac}       % compact symbols for 1/2, etc.

\newtheorem{theorem}{Theorem} % 定义一个定理环境
\renewenvironment{proof}{{\bf Proof.}}{\qed}

\newcommand{\True}{\mathbf{true}}
\newcommand{\False}{\mathbf{false}}

\usepackage{newfloat}
\usepackage{listings}
\DeclareCaptionStyle{ruled}{labelfont=normalfont,labelsep=colon,strut=off} % DO NOT CHANGE THIS
\floatstyle{ruled}
\newfloat{listing}{tb}{lst}{}
\floatname{listing}{Listing}
\newcounter{checksubsection}
\newcounter{checkitem}[checksubsection]

\title{Autonomous Concept Drift Threshold Determination}
\author{
    Pengqian Lu\textsuperscript{\rm 1},
    Jie Lu\textsuperscript{\rm 1}\thanks{$^\dagger$Corresponding author.},
    Anjin Liu\textsuperscript{\rm 1},
    En Yu\textsuperscript{\rm 1}, 
    Guangquan Zhang\textsuperscript{\rm 1}
}

\affiliations{
    \textsuperscript{\rm 1}Australian Artificial Intelligence Institute (AAII),\\
    University of Technology Sydney, Ultimo, NSW 2007, Australia\\
    \{Pengqian.Lu@student., Jie.Lu@, Anjin.Liu@, En.Yu-1@, Guangquan.Zhang@\}uts.edu.au
}

\begin{document}

\maketitle

\begin{abstract}
Existing drift detection methods focus on designing sensitive test statistics. They treat the detection threshold as a fixed hyperparameter, set once to balance false alarms and late detections, and applied uniformly across all datasets and over time.
However, maintaining model performance is the key objective from the perspective of machine learning, and we observe that model performance is highly sensitive to this threshold.
This observation inspires us to investigate whether a dynamic threshold could be provably better. In this paper, we prove that a threshold that adapts over time can outperform any single fixed threshold. The main idea of the proof is that a dynamic strategy, constructed by combining the best threshold from each individual data segment, is guaranteed to outperform any single threshold that apply to all segments. Based on the theorem, we propose a Dynamic Threshold Determination algorithm. It enhances existing drift detection frameworks with a novel comparison phase to inform how the threshold should be adjusted. Extensive experiments on a wide range of synthetic and real-world datasets, including both image and tabular data, validate that our approach substantially enhances the performance of state-of-the-art drift detectors.

% 直接进入主题，少用连接词

\end{abstract}
\begin{links}
\link{Code}{https://github.com/AAII-DeSI/concept-drift-RocStone/tree/main/AAAI2026-DTD} 
\end{links}

\section{Introduction}

In many applications, including network intrusion detection \cite{park2018network} and solar forecasting \cite{wojtkiewicz2018concept}, data is generated as a continuous stream whose underlying distribution is non-stationary and may change over time \cite{survey}. This phenomenon is termed as \emph{concept drift}, which can significantly degrade model performance.
A user-defined threshold is central to handling this drift. Typically, a hypothesis test statistic is monitored: when it crosses this threshold, a drift is signaled. This triggers an adaptation procedure, such as retraining \cite{ddm,EDDM}, to update the model for the new concept.

% 【重点前置，看第一句就知道整段要讲什么】
Traditionally, threshold selection has been seen as a trade-off: a lenient threshold risks delayed detection (leaving the model mismatched with new data), while a stricter threshold risks frequent false alarms (leading to excessive adaption and possible drops in accuracy). This view of the threshold is reflected in the design of many drift detectors \cite{HDDM,adwin}.

A recent study shows that by calibrating thresholds for sensitivity, different statistical tests can achieve similar model performance, i.e., online prediction accuracy \cite{liu2022concept}. This inspired us to question the conventional view of the threshold as merely a tool for balancing statistical trade-offs. \textbf{Can we achieve better model performance by dynamically adjusting the threshold?}

\begin{figure}[t]
    \centering
    \includegraphics[width=\linewidth]{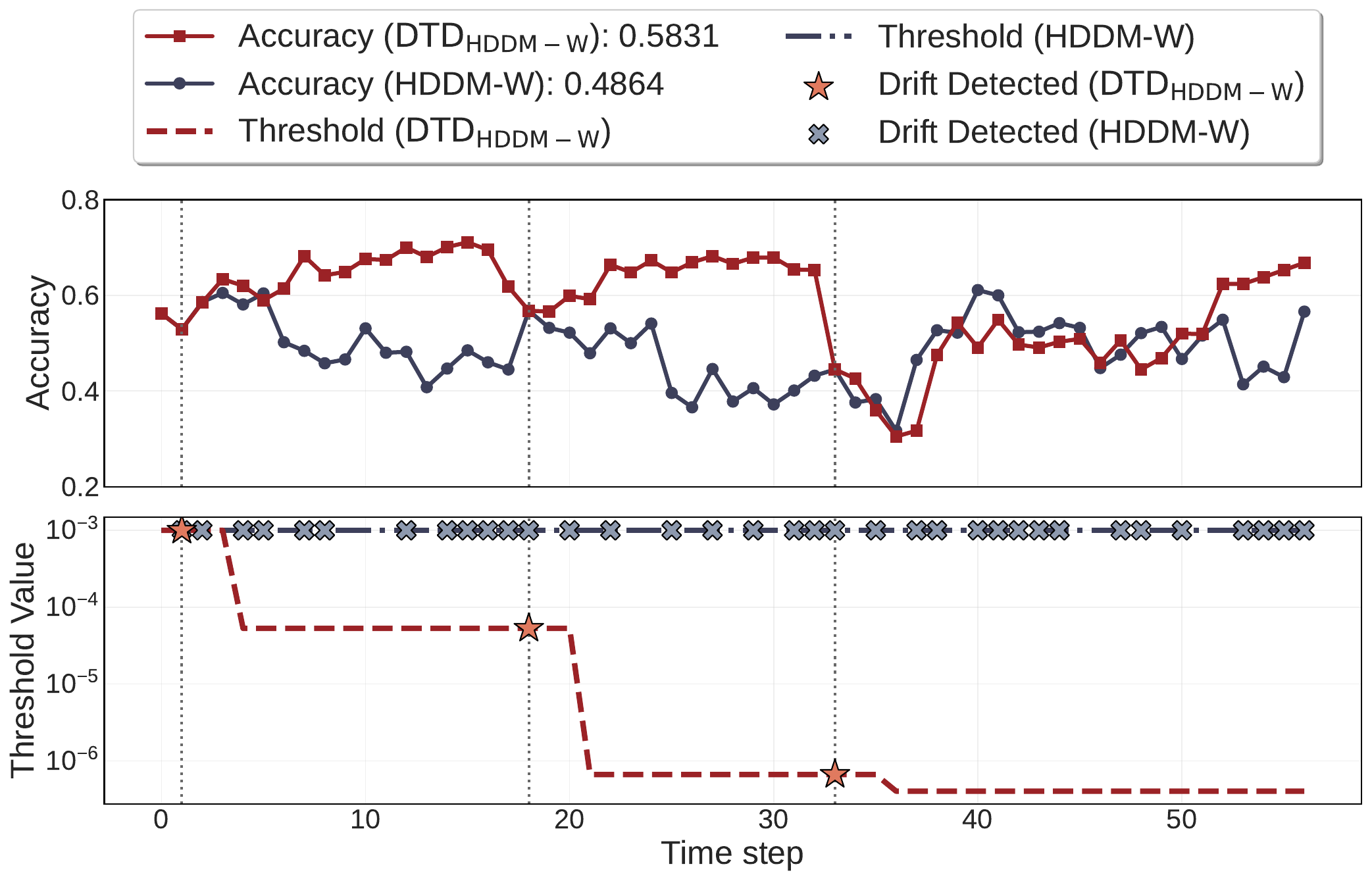}
    \caption{
        A case study on the {Airline} dataset shows the classic {HDDM-W} detector is overly sensitive, raising 36 alarms for a low 48.64\% accuracy. 
        By applying our {DTD} algorithm, the enhanced {$\text{DTD}_{\text{HDDM-W}}$} detector dynamically adapts its threshold and trigger only three alarms, significantly boosting mean accuracy to 58.31\%.
    }
    \label{fig:case-study}
\end{figure}

In this paper, we answer this question affirmatively. We first prove that {simply balancing delayed detection against false alarms does not guarantee optimal performance under concept drift}. We then prove that no single fixed threshold can be universally optimal. Finally, we prove that {dynamic} threshold can significantly outperform any fixed threshold.

Motivated by these insights, we propose a {dynamic threshold determination algorithm}  (DTD) to adjust the threshold in response to the current data, the model’s state, and the chosen adaptation method. 
The main idea is that when a drift is detected, our algorithm runs three models in parallel for several time steps to test three hypotheses: that the detection was too late, correctly timed, or a false alarm. The model with best performance during this comparison directly informs how to adjust the threshold for future detections.
By placing threshold design at the forefront of drift handling, our work offers a concrete direction for practitioners seeking to bridge the gap between theoretical ideals and the practical challenges of real-world data streams. 
As shown in Figure~\ref{fig:case-study}, we provide a case study on the real world Airline dataset. This case study highlights the effectiveness of our propose algorithm. By dynamically adjusting the detection threshold, our method reduced 36 drift alarms to just 3, boosting mean accuracy from a low 48.64\% to 58.31\%.
The contributions of the paper are summarized below.

\begin{enumerate}
    \item We first prove that the conventional goal of balancing detection tradeoff does not guarantee optimal model performance. We argue for shifting the focus from statistical trade-off to maintaining model performance.

    \item We prove that no single, fixed threshold can be universally optimal. Furthermore, we prove that a dynamic threshold strategy is strictly superior to any static one, providing a firm theoretical basis for online adaptation.

    \item Based on these insights, we propose the Dynamic Threshold Determination (DTD) algorithm. DTD introduces a novel comparison phase after a drift signal, using the performance of candidate models to intelligently adjust the threshold for future use.

    \item We conduct extensive experiments on a wide range of synthetic and real-world datasets. The results demonstrate that our DTD algorithm enhances the performance of drift detectors in online data stream scenarios.
\end{enumerate}

\section{Related Work}

Concept drift detection is typically addressed through two dominant paradigms: those that analyze the statistical properties of the data stream, and those that monitor model performance. One family of techniques quantifies dissimilarity between data samples, for instance, through statistical density estimation \cite{song2007statistical}. Histograms are a popular tool for representing distributions, especially in high-dimensional feature spaces \cite{liu2017regional}, with innovations including the use of hierarchical structures \cite{boracchi2018quanttree} and dynamically adjustable binning strategies \cite{yonekawa2022riden}. Alternative partitioning methods like QuadTree \cite{coelho2023concept} and K-means clustering \cite{eikmeans} have also been utilized. Some approaches extend beyond direct statistical comparisons by incorporating contextual factors \cite{lu2018structural}, using context-aware functions like CoDiTE \cite{cobb2022context, CoDiTE}, or forecasting future distributions \cite{ddgda}. Some recent works focus on concept drift adaption on multi-stream  \cite{yu2024online,yu2025drift} or multi-modal LLM setting \cite{yang2025walking}.

% A primary drawback of these distribution-focused strategies is their potential  for high computational cost on high-dimensional data streams \cite{souza2021efficient}. Notably, the concept of building histograms from prediction uncertainty to detect drift remains an unexplored area in the literature.

The second major paradigm, based on error rates, is often favored for its computational efficiency. Well-established detectors such as DDM \cite{ddm}, EDDM \cite{EDDM}, and HDDM \cite{HDDM} function by monitoring fluctuations in the model's error rate. Refinements to this approach include adaptive window resizing \cite{adwin} and forgetting mechanisms that dynamically weight classifiers \cite{iwe}. More recent strategies have incorporated Gaussian Mixture Models or Fourier transform for comparing data windows \cite{yu2024fuzzy,yu2025learning}, implemented reactive states that activate upon alarm detection \cite{driftsurf}, detecting concept drift based on fine-grained error rate \cite{aaai}, monitoring the change of loss value \cite{zhou2, zhou1}, or setting a threshold for true positive rate \cite{xiaoyu}.

Existing drift detectors often rely on predefined thresholds or p-values to manage the trade-off between false alarms and detection speed. We will prove that such fixed settings prevent optimal performance in the next section. Importantly, we also demonstrate that dynamic thresholds are superior to static ones. We believe that this is the first study to introduce a strategy for automatically adjusting thresholds to maximize a model's overall effectiveness when data changes.

\section{Methodology}
\subsection{Problem Setup}

Let us denote a stream as $D$, which includes some labeled samples 
$\{(x_t, y_t)\}_{t = 1}^{T}$. Here, each $x_t \in \mathcal{X}$ 
is the instance collected at time $t$, and $y_t \in \mathcal{Y}$ 
is its corresponding label. The size of stream $T$ may be large or potentially unbounded. If the stream is collected in chunks, we denote the chunk collect at time $t$ as $C_t=\{(x_i,y_i)|i\in [1, |C_t|]\}$ where $|C_t|$ is the size of the chunk.  The joint distributions on $\mathcal{X} \times \mathcal{Y}$ is denoted as $\{P_t\}_{t=1}^T$ , where $P_t$ generates $(x_t, y_t)$ at time $t$. If $P_t$ remains identical for all $t$, there is {no concept drift}. Otherwise, if there is at least one time step $t$ such that $P_t \neq P_{t+1}$, we claim the concept drift occurs at time $t+1$. We then consider $P_t$ as the {old concept} and $P_{t+1}$ as the {new concept}.

To detect whether a concept drift occurs at time $t$, 
we define a data window $\Omega_t$ as 
$$
\Omega_t \;=\; \{\, (x_k,y_k) \,\mid\, k \in [t-W+1,\dots,t] \},
$$
where $W$ is the size of the window and $t \ge W$. 
Drift detectors often split $\Omega_t$ into several sub-windows,
compare distributions, or apply hypothesis tests to detect drift. Let $S_t$ denote the test statistic computed on $\Omega_t$, which can be considered as a function of $\Omega_t$:
$$
S_t \;=\; f\bigl(\Omega_t\bigr),
$$
where $f(\cdot)$ is any statistic designed to signal 
a possible distribution change (e.g., an error-increase measure).

\paragraph{Drift Alarm.}
Given a fixed threshold $\theta$ at each time $t$, we pose 
a hypothesis test:
$$
H_0: \text{No drift at time $t$} 
\quad \text{vs.} \quad
H_1: \text{Drift at time $t$}.
$$
If $S_t > \theta$, we reject null hypothesis and  raise a drift detection alarm at $t$. 
Different methods define $f(\cdot)$  differently, 
but almost all compare a final statistic $S_t$ to $\theta$.

\paragraph{False Alarm.}
A {false alarm} arises when no actual drift is present 
but an alarm is raised. Formally, we define the probability of an false alarm at $t$ as
$$
\Pr[\text{false alarm at } t] 
= \Pr\bigl(S_t > \theta \,\mid\, P_t = P_{t-1}\bigr).
$$
In practice, the design of drift detector seeks to keep this probability low to avoid frequent drift adaption.

\paragraph{Detection Delay.}
Assume a true concept drift occurs at time $t$ (i.e., $P_t \neq P_{t-1}$). The {detection delay}, denoted by $\Delta(t)$, is the number of time steps required for the detector to raise an alarm. Formally, it is the smallest non-negative integer $d$ such that the test statistic $S_{t+d}$ exceeds a predefined threshold $\theta$. A delay of $\Delta(t)=0$ indicates an immediate detection. The probability of a specific delay $d$ is given by:
\begin{align*}
\Pr[\Delta(t) = d] ={}& \Pr\bigl(S_{t}\le \theta, \dots, S_{t+d-1} \le \theta, \nonumber \\
                      & \phantom{=\Pr\bigl(} S_{t+d} > \theta \;\mid\; P_t \neq P_{t-1}\bigr).
\end{align*}
Any $d>0$ constitutes a {delayed detection}, as the alarm is raised only after additional data points have been observed.

\paragraph{Perfect Detection.} Without loss of generality, we assume the statistic $S_t$ is a measure of dissimilarity and a drift alarm will be raised when $S_t > \theta$. The threshold $\theta$ thus governs the critical trade-off between detection delay and false alarm. A lower threshold enhances sensitivity, enabling fast detection but at the cost of more frequent false alarms. A higher threshold ensures robustness against false alarms but at the expense of detection latency for actual drifts.
We define a perfect, idealized detector as one that, for all $t$, simultaneously achieves zero false alarms and zero delay. Formally, it is a detector satisfies
$$
\Pr\bigl[\text{false alarm at } t\bigr] = 0
\quad\text{and}\quad
\Pr\bigl[\Delta(t) = 0\bigr] = 1.
$$
% This ideal, however, is precluded by the statistical uncertainty arising from the noise and finite samples. Therefore, the trade-off is inescapable: No single $\theta$ can eliminate both error types simultaneously.

% This follows your Problem Setup section.
\paragraph{Model Performance}
Drift adaptation is triggered when a test statistic $S_t$ crosses a threshold. The rule for setting this threshold is the {threshold strategy}, denoted by $\boldsymbol{\theta}$. A strategy can be:
1) A fixed threshold $\boldsymbol{\theta} = \theta_{\mathrm{const}}$.
2) A sequence of thresholds varying over time $\boldsymbol{\theta} = \{\theta_t\}_{t=1}^T$.
The performance of a strategy $\boldsymbol{\theta}$ on a stream $D$ is its online accuracy, defined using a 0-1 loss function $\ell(\cdot, \cdot)$:
\[
A(\boldsymbol{\theta}; D) = \frac{1}{T} \sum_{t=1}^{T} (1 - \ell(\hat{y}_t, y_t)).
\]
Note that the predictions $\{\hat{y}_t\}$ depend on $\boldsymbol{\theta}$, as it determines when the model conducts drift adaption. For brevity, we denote the performance on a stream $D$ as $A(\cdot; D)$.

\subsection{Theoretical Analysis}

This section establishes the theoretical foundation that motivates the development of dynamic thresholding algorithms. We present three theorems that formalize the limitations of conventional fixed-threshold approaches and prove the superiority of a dynamic strategy. {Due to space constraints, all proofs are deferred to the Appendix.}
First, we challenge the notion that perfect detection is always optimal.

\begin{theorem}[Perfect Detection May Not Be Optimal]
\label{thm:perf_vs_delayed}
Perfect detection of concept drift may fail to yield optimal model performance in a streaming setting.
\end{theorem}

This implies that even a statistically perfect detection, with zero delay and no false alarms, does not necessarily maximize model performance. 
For instance, detecting a very subtle drift might trigger an unnecessary adaptation, causing the model to forget valuable prior knowledge and ultimately harming its overall accuracy. 
This insight suggests that drift detection should focus more on preserving model performance rather than just achieving statistical perfection. Our second theorem challenges the notion that the threshold should be treated as a predefined, fixed value.

\begin{theorem}[No Single Threshold is Universally Optimal]
\label{thm:no_universal_threshold}
No single drift-detection threshold guarantees optimal performance on every dataset, model, and adaptation method.
\end{theorem}

These limitations motivate our final theorem, which establishes the formal superiority of a dynamic approach. 

\begin{theorem}[Dynamic Thresholds Outperform Stationary Thresholds]
Consider a data stream $D$. Let $\Theta_{\mathrm{const}}$ be the set of all stationary thresholds and $\Theta_{\mathrm{dyn}}$ be the set of all dynamic-threshold strategies. Let $A(\cdot; D)$ be the model performance on a stream $D$. Then:
\[
\max_{\{\theta_t\}\,\in\,\Theta_{\mathrm{dyn}}} \,A(\{\theta_t\};\,D)
\;\;\ge\;\;
\max_{\theta\,\in\,\Theta_{\mathrm{const}}} \,A(\theta;\,D).
\]
\end{theorem}

This result provides the theoretical justification for designing algorithms that adapt the detection threshold over time, which is the core contribution of this work.

\subsection{Dynamic Threshold Determination Algorithm}
    
Our proposed Dynamic Threshold Determination (DTD) algorithm, detailed in Algorithm~\ref{algo-1}, adaptively adjusts the threshold of a concept drift detector. At time step $t$, its core mechanism will be triggered if the detector's statistic $S_t$ exceeds the current threshold $\theta$. Instead of immediately conduct drift adaption, DTD initiates three candidate models enter a comparison phase. The threshold is then adjusted based on the relative performance of these models.

\begin{enumerate}
    \item \textbf{Early Drift Model} (\texttt{EDM}): This model represents an aggressive strategy, assuming that the drift was detectable \textit{before} the current time step $t$. Consequently, it initiates adaptation based on the data collected at last time step ${t-1}$. 
    If \texttt{EDM}  performs best among all of the candidate models, it suggests the initial detection was delayed. DTD then sets the threshold to the detector statistic from the previous time step, $\theta \leftarrow S_{i-1}$, to enhance sensitivity for the earlier detection of future drifts.
    
    \item \textbf{Reactive Drift Model} (\texttt{RDM}): This model embodies a standard strategy and assumes the current sensitivity is appropriate. If \texttt{RDM} excels, it indicates that the detection timing and current threshold are appropriate. Accordingly, $\theta$ remains unchanged.

    \item \textbf{Previous Model} (\texttt{PM}): This model holds the assumption that the drift signal at $t$ is a false alarm. It thus refuse to conduct drift adaption at time step $t$. If \texttt{PM} demonstrates superior performance, it implies the system was overly sensitive and the signal at $t$ was likely a false alarm. Therefore, DTD increases the threshold to $\theta \leftarrow S_i+\eta$, where $\eta$ is a small positive constant. This adjustment aims to prevent similar false signals from triggering drift detections in the future.
\end{enumerate}

\begin{algorithm}[t]
\caption{Dynamic Threshold Determination Algorithm}
\begin{algorithmic}[1]
\STATE \textbf{Input:} Data Stream $\mathcal{D} = \{ C_t \}$; Drift Detector $\psi$; Initial $\theta_0$; Initial Model $M_0$; Leading candidate model $M_l = \texttt{RDM}$; Prediction model $M = M_0$; Threshold $\theta = \theta_0$; List of accuracy $\Lambda = []$; Comparison phase flag $\Gamma = \False$; Comparison steps $k = K$; Last model $M'=M_0$; A extreme small constant $\eta$.
\STATE \textbf{Output:} $\text{Avg}(\Lambda)$. 
\STATE List of candidate models $\mathcal{M} = \emptyset$.
\STATE List of accuracy of candidate models $\Psi=\emptyset$.
\STATE List of candidate drift detectors $\Pi = \emptyset$.
\FOR{each chunk $C_t \in \mathcal{D}$}
    \IF{$\Gamma=\False$}
        \STATE $a_i, S_{t} = \text{Evaluate}(M, C_t, \psi)$
        \IF{$S_{t} > \theta$} 
        \STATE $\mathcal{M}, \Pi, \Psi = \text{CreateCandidates}(M, M', C_t, C_{t-1},$
        \item[] \hspace*{\algorithmicindent} $a_t, S_t, S_{t-1})$ \# See Appendix
        \STATE $\Gamma = \True$; $k = K$; $M_l = \texttt{RDM}$
        \ENDIF
         \STATE $M'=\text{copy}(M)$; 
         \STATE $\text{Train}(M, C_t)$ \# If continual training
    \ELSE 
        \STATE $A = \text{EvalCandidates} (\mathcal{M}, C_t, \Pi)$ \# See Appendix
        \STATE $k = k - 1$; $a_t = A[M_l]$; $M_l = \text{arg max}_{\texttt{name}} [A]$
        \IF{$k = 0$} 
            % \STATE $M_{win} = \text{DetermineWinningModel}(\Pi, \mathcal{M})$
            \STATE $\Pi=\{\texttt{EDM}: \text{Avg}(\Pi[\texttt{EDM}]), \texttt{PM}: \text{Avg}(\Pi[\texttt{PM}]),$ 
            \item[] \hspace*{\algorithmicindent} $ \texttt{RDM}: \text{Avg}(\Pi[\texttt{RDM}])\}$
            \STATE $M_l=\arg\max_{name}[\Pi]$
            \STATE $M=\mathcal{M}[M_l]$; $\psi=\Psi[M_l]$
            \STATE $\theta=$ threshold of $\Psi[M_l]$;
            \STATE $\Gamma = \False$; $\mathcal{M} = \emptyset$; $\Pi = \emptyset$; $M_l = \texttt{RDM}$;
        \ENDIF
    \ENDIF
    \STATE Add $a_t$ to $\Lambda$
\ENDFOR
\end{algorithmic}
\label{algo-1}
\end{algorithm}

Specifically, the algorithm operates in two primary phases: a {Normal Operation Phase} and a {Comparison Phase}.

\paragraph{Normal Operation Phase.}
During this phase, the system employs a primary predictive model $M$ to process incoming data chunks $C_t$. For each chunk, $M$ is evaluated, yielding an accuracy $a_t$, and a detector-specific statistic $S_t$ is computed. If this statistic $S_t$ exceeds the current threshold $\theta$ (or falls below it, depending on the nature of the detector), a potential concept drift is signaled. At this time, the system records $S_t$, the statistic from the previous chunk $S_{t-1}$, and instantiates three candidate models for the comparison phase.

As detailed in Appendix Algorithm 2, three distinct candidate models are instantiated, each representing a different hypothesis regarding the suspected drift:
\begin{enumerate}
    \item \texttt{EDM} assumes that the true drift occurred at the preceding chunk $C_{t-1}$. It is therefore constructed by adapting the predictive model at last time step $M'$ using data $C_{t-1}$. 
    % Subsequently, \texttt{EDM} is evaluated and incrementally trained on the current chunk $C_i$.
    \item \texttt{RDM} is constructed by adapting the prediction model $M$ using data from the current chunk $t_i$.
    \item \texttt{PM} is a direct copy of the primary model $M$, embodying the hypothesis that the drift signal was a false alarm.
\end{enumerate}
To monitor these models during the comparison phase, DTD also initializes three corresponding drift detectors. The threshold for the detector associated with \texttt{EDM} is set to $S_{t-1}$. The threshold remains unchanged for \texttt{RDM}. The drift detector for \texttt{PM} is a copy of the primary drift detector used in the normal operation phase with the threshold set as $S_t+\eta$.

\paragraph{Comparison Phase.}
Upon a drift detection at time step $t$, DTD initiates the comparison phase to ascertain the nature of the detected change and to inform the threshold adjustment. This phase spans $K$ subsequent data chunks, during which the candidate models are evaluated and incrementally updated using these new chunks if the training strategy is set as continual learning. The three corresponding drift detectors continuously monitor the performance of respective candidate models. If a detector signals a drift for its associated model, that model undergoes adaptation accordingly.

Upon completion of the $K$-chunk comparison phase, DTD compares the accumulated performance of the candidate models during this period as described in Appendix Algorithm 3. The winning model and its corresponding drift detector are then selected to become the new primary predictive model and its associated drift detector for the subsequent normal operation phase.

\paragraph{Time Complexity Analysis.}
Assuming the base detector's complexity is $O(n)$, our framework is also $O(n)$ during normal operation. The complexity temporarily increases to $O(3n)$ during the K-step comparison phase due to maintaining three parallel models. In the extreme worst case (i.e., the stream constantly triggers the comparison phase), the overall complexity is $O(3n)$. Crucially, this demonstrates that the overhead is a linear increase relative to the base detector, not an exponential one.

\begin{table*}[!t]
\centering
\small
\setlength{\tabcolsep}{1mm} % 调整列间距

% The two parts of the table are placed side-by-side
% Part 1
\begin{tabular}{ @{} ll cc cc cc cc @{} }
\toprule
\multirow{2}{*}{\textbf{Dataset}} & \multirow{2}{*}{\textbf{\begin{tabular}[c]{@{}c@{}}Training\end{tabular}}} & \multicolumn{2}{c}{\textbf{KSWIN}} & \multicolumn{2}{c}{\textbf{DDM}} & \multicolumn{2}{c}{\textbf{PH}} & \multicolumn{2}{c}{\textbf{HDDM-A}} \\
\cmidrule(lr){3-4} \cmidrule(lr){5-6} \cmidrule(lr){7-8} \cmidrule(lr){9-10}
 &  & Baseline & $\text{DTD}_{\text{KSWIN}}$ & Baseline & $\text{DTD}_{\text{DDM}}$ & Baseline & $\text{DTD}_{\text{PH}}$ & Baseline & $\text{DTD}_{\text{HDDM\text{-}A}}$ \\ \midrule
\multirow{2}{*}{Airline} & \small continual & 50.21$\pm$1.95 & \textbf{57.29$\pm$4.44} & 52.94$\pm$0.00 & \textbf{53.60$\pm$0.00} & 49.35$\pm$0.00 & \textbf{52.69$\pm$0.00} & 52.80$\pm$0.00 & \textbf{52.98$\pm$0.00} \\
 & \small sporadic & 48.69$\pm$0.95 & \textbf{53.49$\pm$2.36} & \textbf{52.43$\pm$0.00} & 51.78$\pm$0.00 & 49.02$\pm$0.00 & \textbf{51.19$\pm$0.00} & \textbf{55.62$\pm$0.00} & 52.17$\pm$0.00 \\
\multirow{2}{*}{Elec2} & \small continual & 67.85$\pm$0.24 & \textbf{69.26$\pm$0.54} & 67.75$\pm$0.00 & \textbf{71.83$\pm$0.00} & 70.12$\pm$0.00 & \textbf{71.28$\pm$0.00} & 67.73$\pm$0.00 & \textbf{70.19$\pm$0.00} \\
 & \small sporadic & 67.59$\pm$0.17 & \textbf{68.21$\pm$0.34} & \textbf{67.60$\pm$0.00} & 66.44$\pm$0.00 & 70.04$\pm$0.00 & \textbf{70.48$\pm$0.00} & 67.73$\pm$0.00 & \textbf{68.23$\pm$0.00} \\
\multirow{2}{*}{PS} & \small continual & \textbf{71.21$\pm$0.13} & 71.11$\pm$0.22 & 69.63$\pm$0.00 & \textbf{71.88$\pm$0.00} & 70.36$\pm$0.00 & \textbf{71.88$\pm$0.00} & 70.87$\pm$0.00 & \textbf{72.01$\pm$0.00} \\
 & \small sporadic & 68.57$\pm$1.85 & \textbf{70.98$\pm$0.35} & 67.52$\pm$0.00 & \textbf{70.74$\pm$0.00} & 68.67$\pm$0.00 & \textbf{70.12$\pm$0.00} & 71.24$\pm$0.00 & \textbf{72.09$\pm$0.00} \\
\multirow{2}{*}{SEA0} & \small continual & 91.03$\pm$0.96 & \textbf{91.66$\pm$1.08} & 94.03$\pm$0.56 & \textbf{94.75$\pm$0.51} & 94.35$\pm$0.31 & \textbf{94.84$\pm$0.23} & 94.27$\pm$0.36 & \textbf{94.75$\pm$0.31} \\
 & \small sporadic & 89.96$\pm$2.35 & \textbf{90.50$\pm$2.11} & 93.49$\pm$0.85 & \textbf{94.21$\pm$0.66} & 94.01$\pm$0.52 & \textbf{94.60$\pm$0.24} & 93.94$\pm$0.40 & \textbf{94.51$\pm$0.29} \\
\multirow{2}{*}{SEA10} & \small continual & 84.20$\pm$0.66 & \textbf{85.05$\pm$0.99} & 85.28$\pm$0.99 & \textbf{87.08$\pm$0.41} & 87.14$\pm$0.18 & \textbf{87.61$\pm$0.18} & 86.84$\pm$0.23 & \textbf{87.27$\pm$0.28} \\
 & \small sporadic & 83.18$\pm$1.32 & \textbf{84.92$\pm$1.62} & 84.79$\pm$1.01 & \textbf{86.35$\pm$0.70} & 86.56$\pm$0.27 & \textbf{87.08$\pm$0.28} & 86.07$\pm$0.62 & \textbf{86.75$\pm$0.41} \\
\multirow{2}{*}{SEA20} & \small continual & 76.12$\pm$0.60 & \textbf{77.13$\pm$0.61} & 76.24$\pm$0.63 & \textbf{77.36$\pm$0.77} & 77.80$\pm$0.20 & \textbf{78.18$\pm$0.20} & 77.50$\pm$0.27 & \textbf{77.85$\pm$0.33} \\
 & \small sporadic & 74.81$\pm$1.00 & \textbf{76.94$\pm$0.70} & 74.88$\pm$1.28 & \textbf{76.76$\pm$0.75} & 76.84$\pm$0.51 & \textbf{77.61$\pm$0.27} & 76.32$\pm$0.59 & \textbf{77.14$\pm$0.58} \\
\multirow{2}{*}{Sine} & \small continual & 81.73$\pm$1.01 & \textbf{82.50$\pm$1.24} & 82.19$\pm$1.44 & \textbf{83.53$\pm$1.28} & 83.55$\pm$1.36 & \textbf{84.39$\pm$1.17} & 83.09$\pm$1.26 & \textbf{83.97$\pm$1.25} \\
 & \small sporadic & 80.84$\pm$1.81 & \textbf{82.06$\pm$1.40} & 81.05$\pm$3.81 & \textbf{83.59$\pm$2.15} & 83.16$\pm$2.23 & \textbf{84.50$\pm$1.30} & 83.28$\pm$1.27 & \textbf{83.36$\pm$2.63} \\
\multirow{2}{*}{Mixed} & \small continual & 83.87$\pm$0.11 & \textbf{84.23$\pm$0.11} & 83.80$\pm$0.14 & \textbf{84.21$\pm$0.11} & 83.87$\pm$0.11 & \textbf{84.23$\pm$0.11} & 83.87$\pm$0.11 & \textbf{84.23$\pm$0.11} \\
 & \small sporadic & 83.51$\pm$0.24 & \textbf{83.92$\pm$0.20} & 82.69$\pm$2.21 & \textbf{83.82$\pm$0.59} & 83.51$\pm$0.24 & \textbf{83.92$\pm$0.20} & 83.52$\pm$0.24 & \textbf{83.92$\pm$0.20} \\
\bottomrule
\end{tabular}
\quad % 两个表格之间的间隔
% Part 2
\begin{tabular}{ @{} ll cc cc cc cc @{} }
\toprule
\multirow{2}{*}{\textbf{Dataset}} & \multirow{2}{*}{\textbf{\begin{tabular}[c]{@{}c@{}}Training\end{tabular}}} & \multicolumn{2}{c}{\textbf{HDDM-W}} & \multicolumn{2}{c}{\textbf{PUDD-1}} & \multicolumn{2}{c}{\textbf{PUDD-3}} & \multicolumn{2}{c}{\textbf{PUDD-5}} \\
\cmidrule(lr){3-4} \cmidrule(lr){5-6} \cmidrule(lr){7-8} \cmidrule(lr){9-10}
 &  & Baseline & $\text{DTD}_{\text{HDDM\text{-}W}}$ & Baseline & $\text{DTD}_{\text{PUDD\text{-}1}}$ & Baseline & $\text{DTD}_{\text{PUDD\text{-}3}}$ & Baseline & $\text{DTD}_{\text{PUDD\text{-}5}}$ \\ \midrule
\multirow{2}{*}{Airline} & \small continual & 48.66$\pm$0.00 & \textbf{58.31$\pm$0.00} & \textbf{53.57$\pm$0.27} & 53.11$\pm$0.75 & 53.03$\pm$0.29 & \textbf{53.71$\pm$0.42} & \textbf{52.16$\pm$0.88} & 51.65$\pm$0.12 \\
 & \small sporadic & 48.62$\pm$0.00 & \textbf{49.88$\pm$0.00} & \textbf{51.05$\pm$0.01} & 50.66$\pm$0.19 & 49.45$\pm$0.43 & \textbf{52.77$\pm$0.09} & \textbf{54.37$\pm$0.77} & 53.22$\pm$0.54 \\
\multirow{2}{*}{Elec2} & \small continual & 67.73$\pm$0.00 & \textbf{70.11$\pm$0.00} & 70.85$\pm$0.52 & \textbf{72.12$\pm$0.14} & 70.85$\pm$0.50 & \textbf{72.16$\pm$0.20} & 70.69$\pm$0.77 & \textbf{72.24$\pm$0.25} \\
 & \small sporadic & 67.73$\pm$0.00 & \textbf{67.95$\pm$0.00} & 62.76$\pm$0.78 & \textbf{69.04$\pm$0.15} & 59.32$\pm$0.75 & \textbf{69.04$\pm$0.15} & 59.44$\pm$0.85 & \textbf{69.04$\pm$0.15} \\
\multirow{2}{*}{PS} & \small continual & 71.06$\pm$0.00 & \textbf{71.90$\pm$0.00} & 71.88$\pm$0.76 & \textbf{71.98$\pm$0.04} & 71.59$\pm$0.28 & \textbf{71.98$\pm$0.04} & 71.59$\pm$0.83 & \textbf{71.80$\pm$0.00} \\
 & \small sporadic & 69.53$\pm$0.00 & \textbf{70.04$\pm$0.00} & \textbf{71.13$\pm$0.69} & 70.65$\pm$0.00 & \textbf{71.20$\pm$0.49} & 70.65$\pm$0.00 & 70.40$\pm$0.10 & \textbf{70.52$\pm$0.00} \\
\multirow{2}{*}{SEA0} & \small continual & 91.90$\pm$1.07 & \textbf{92.73$\pm$1.07} & 94.61$\pm$0.07 & \textbf{94.96$\pm$0.29} & 94.81$\pm$0.58 & \textbf{94.97$\pm$0.20} & 94.85$\pm$0.76 & \textbf{94.97$\pm$0.19} \\
 & \small sporadic & 91.63$\pm$0.97 & \textbf{92.53$\pm$1.05} & 94.25$\pm$0.73 & \textbf{94.66$\pm$0.26} & 94.60$\pm$0.56 & \textbf{94.65$\pm$0.25} & \textbf{94.62$\pm$0.22} & \textbf{94.62$\pm$0.28} \\
\multirow{2}{*}{SEA10} & \small continual & 85.78$\pm$0.72 & \textbf{86.75$\pm$0.60} & 87.24$\pm$0.94 & \textbf{87.67$\pm$0.16} & 87.28$\pm$0.13 & \textbf{87.67$\pm$0.15} & 87.18$\pm$0.18 & \textbf{87.60$\pm$0.19} \\
 & \small sporadic & 85.31$\pm$0.50 & \textbf{86.68$\pm$0.45} & 86.53$\pm$0.07 & \textbf{87.22$\pm$0.25} & 86.61$\pm$0.04 & \textbf{87.08$\pm$0.32} & 86.55$\pm$0.41 & \textbf{87.01$\pm$0.28} \\
\multirow{2}{*}{SEA20} & \small continual & 77.70$\pm$0.23 & \textbf{77.92$\pm$0.25} & 78.08$\pm$0.73 & \textbf{78.33$\pm$0.19} & 77.86$\pm$0.28 & \textbf{78.32$\pm$0.17} & 77.68$\pm$0.85 & \textbf{78.19$\pm$0.22} \\
 & \small sporadic & 76.86$\pm$0.33 & \textbf{77.41$\pm$0.42} & 76.89$\pm$0.34 & \textbf{77.55$\pm$0.36} & 77.02$\pm$0.58 & \textbf{77.48$\pm$0.48} & 76.67$\pm$0.78 & \textbf{77.19$\pm$0.65} \\
\multirow{2}{*}{Sine} & \small continual & 82.46$\pm$1.08 & \textbf{83.56$\pm$1.35} & 83.12$\pm$0.19 & \textbf{84.30$\pm$1.23} & 83.39$\pm$0.35 & \textbf{84.31$\pm$1.10} & 83.43$\pm$0.51 & \textbf{84.15$\pm$1.05} \\
 & \small sporadic & 82.61$\pm$1.87 & \textbf{84.02$\pm$1.40} & 81.48$\pm$0.11 & \textbf{83.61$\pm$2.90} & 83.80$\pm$0.16 & \textbf{84.39$\pm$1.24} & 83.38$\pm$0.82 & \textbf{84.16$\pm$1.18} \\
\multirow{2}{*}{Mixed} & \small continual & 83.87$\pm$0.11 & \textbf{84.23$\pm$0.11} & 82.99$\pm$0.13 & \textbf{83.87$\pm$0.46} & 83.92$\pm$0.41 & \textbf{83.99$\pm$0.38} & 84.12$\pm$0.60 & \textbf{84.13$\pm$0.23} \\
 & \small sporadic & 83.51$\pm$0.24 & \textbf{83.92$\pm$0.20} & 79.23$\pm$0.67 & \textbf{82.49$\pm$2.49} & 83.58$\pm$0.19 & \textbf{83.93$\pm$0.20} & \textbf{83.96$\pm$0.30} & 83.92$\pm$0.20 \\
\bottomrule
\end{tabular}

\caption{Performance comparison with classic drift detector and PUDD using the \textbf{GNB} classifier. We compare each baseline against our proposed method $\text{DTD}_{\text{Baseline}}$. The results are presented as mean accuracy (\%) $\pm$ standard deviation (multiplied by 100 for space efficiency). The best performance in each pair is highlighted in \textbf{bold}. PS is short for powersupply.}
\label{tab:gnb_results}
\end{table*}

\begin{table*}[!t]
\centering
\small
\setlength{\tabcolsep}{1mm} % 调整列间距

% The two parts of the table are placed side-by-side
% Part 1
\begin{tabular}{ @{} ll cc cc cc cc @{} }
\toprule
\multirow{2}{*}{\textbf{Dataset}} & \multirow{2}{*}{\textbf{\begin{tabular}[c]{@{}c@{}}Training\end{tabular}}} & \multicolumn{2}{c}{\textbf{KSWIN}} & \multicolumn{2}{c}{\textbf{DDM}} & \multicolumn{2}{c}{\textbf{PH}} & \multicolumn{2}{c}{\textbf{HDDM-A}} \\
\cmidrule(lr){3-4} \cmidrule(lr){5-6} \cmidrule(lr){7-8} \cmidrule(lr){9-10}
 &  & Baseline & $\text{DTD}_{\text{KSWIN}}$ & Baseline & $\text{DTD}_{\text{DDM}}$ & Baseline & $\text{DTD}_{\text{PH}}$ & Baseline & $\text{DTD}_{\text{HDDM\text{-}A}}$ \\ \midrule
\multirow{2}{*}{Airline} & \small continual & 61.05$\pm$2.22 & \textbf{64.36$\pm$0.55} & 61.49$\pm$2.41 & \textbf{65.70$\pm$0.29} & 60.47$\pm$2.24 & \textbf{65.65$\pm$0.29} & 61.72$\pm$2.15 & \textbf{65.43$\pm$0.72} \\
 & \small sporadic & 60.90$\pm$0.65 & \textbf{61.80$\pm$1.38} & 57.88$\pm$1.67 & \textbf{61.07$\pm$0.90} & 60.19$\pm$0.49 & \textbf{61.77$\pm$0.81} & 59.17$\pm$0.79 & \textbf{61.81$\pm$0.81} \\
\multirow{2}{*}{Elec2} & \small continual & 73.45$\pm$1.30 & \textbf{74.17$\pm$0.48} & 73.41$\pm$1.20 & \textbf{76.31$\pm$0.40} & 74.15$\pm$0.84 & \textbf{75.43$\pm$0.56} & 73.29$\pm$1.59 & \textbf{75.77$\pm$0.46} \\
 & \small sporadic & 72.54$\pm$1.21 & \textbf{73.40$\pm$0.47} & \textbf{72.81$\pm$1.32} & 71.99$\pm$0.54 & 72.18$\pm$0.93 & \textbf{72.86$\pm$0.71} & 72.71$\pm$1.42 & \textbf{72.97$\pm$0.58} \\
\multirow{2}{*}{PS} & \small continual & 70.92$\pm$2.03 & \textbf{72.24$\pm$0.18} & 71.34$\pm$0.55 & \textbf{72.26$\pm$0.16} & 71.09$\pm$1.90 & \textbf{72.22$\pm$0.25} & 69.62$\pm$2.74 & \textbf{72.23$\pm$0.17} \\
 & \small sporadic & 67.52$\pm$2.72 & \textbf{71.99$\pm$0.34} & 65.58$\pm$3.74 & \textbf{71.14$\pm$0.38} & 69.07$\pm$0.65 & \textbf{70.18$\pm$0.26} & 68.74$\pm$3.57 & \textbf{71.83$\pm$0.54} \\
\multirow{2}{*}{SEA0} & \small continual & 97.96$\pm$0.20 & \textbf{98.77$\pm$0.10} & 97.19$\pm$0.70 & \textbf{98.73$\pm$0.08} & 97.96$\pm$0.28 & \textbf{98.70$\pm$0.15} & 97.98$\pm$0.22 & \textbf{98.73$\pm$0.11} \\
 & \small sporadic & 91.12$\pm$1.95 & \textbf{92.08$\pm$2.53} & 93.66$\pm$3.83 & \textbf{97.73$\pm$0.44} & 97.04$\pm$0.44 & \textbf{97.87$\pm$0.16} & 96.78$\pm$0.63 & \textbf{97.77$\pm$0.18} \\
\multirow{2}{*}{SEA10} & \small continual & 88.18$\pm$0.28 & \textbf{89.13$\pm$0.09} & 88.23$\pm$0.25 & \textbf{89.06$\pm$0.08} & 87.64$\pm$2.17 & \textbf{88.86$\pm$0.20} & 87.70$\pm$2.88 & \textbf{88.98$\pm$0.16} \\
 & \small sporadic & 82.80$\pm$1.20 & \textbf{85.12$\pm$1.23} & 85.04$\pm$1.28 & \textbf{86.54$\pm$0.47} & 86.79$\pm$0.42 & \textbf{87.08$\pm$0.25} & 86.44$\pm$0.38 & \textbf{86.79$\pm$0.28} \\
\multirow{2}{*}{SEA20} & \small continual & 77.51$\pm$1.47 & \textbf{79.17$\pm$0.14} & 78.03$\pm$0.38 & \textbf{79.16$\pm$0.17} & 77.75$\pm$1.12 & \textbf{78.93$\pm$0.17} & 77.49$\pm$1.56 & \textbf{79.09$\pm$0.12} \\
 & \small sporadic & 74.55$\pm$0.77 & \textbf{76.32$\pm$0.78} & 74.47$\pm$1.27 & \textbf{75.54$\pm$0.93} & 76.56$\pm$0.65 & \textbf{76.88$\pm$0.26} & 75.86$\pm$0.67 & \textbf{76.40$\pm$0.43} \\
\multirow{2}{*}{Sine} & \small continual & 77.74$\pm$9.97 & \textbf{94.14$\pm$0.82} & 87.36$\pm$6.07 & \textbf{93.56$\pm$1.20} & 77.03$\pm$9.75 & \textbf{93.96$\pm$1.02} & 81.34$\pm$9.84 & \textbf{94.39$\pm$0.67} \\
 & \small sporadic & 84.67$\pm$1.72 & \textbf{87.35$\pm$2.04} & 82.70$\pm$4.13 & \textbf{89.02$\pm$5.26} & 88.64$\pm$1.44 & \textbf{91.28$\pm$0.97} & 88.38$\pm$1.60 & \textbf{91.32$\pm$0.95} \\
\multirow{2}{*}{Mixed} & \small continual & 78.72$\pm$9.05 & \textbf{88.85$\pm$0.15} & 84.74$\pm$1.10 & \textbf{89.97$\pm$0.03} & 82.51$\pm$7.07 & \textbf{88.02$\pm$0.24} & 81.42$\pm$8.43 & \textbf{89.99$\pm$0.03} \\
 & \small sporadic & 85.26$\pm$0.51 & \textbf{86.87$\pm$0.11} & 83.95$\pm$2.79 & \textbf{86.07$\pm$2.14} & 85.26$\pm$0.57 & \textbf{86.83$\pm$0.15} & 85.44$\pm$0.47 & \textbf{86.47$\pm$1.58} \\
\bottomrule
\end{tabular}
\quad % 两个表格之间的间隔
% Part 2
\begin{tabular}{ @{} ll cc cc cc cc @{} }
\toprule
\multirow{2}{*}{\textbf{Dataset}} & \multirow{2}{*}{\textbf{\begin{tabular}[c]{@{}c@{}}Training\end{tabular}}} & \multicolumn{2}{c}{\textbf{HDDM-W}} & \multicolumn{2}{c}{\textbf{PUDD-1}} & \multicolumn{2}{c}{\textbf{PUDD-3}} & \multicolumn{2}{c}{\textbf{PUDD-5}} \\
\cmidrule(lr){3-4} \cmidrule(lr){5-6} \cmidrule(lr){7-8} \cmidrule(lr){9-10}
 &  & Baseline & $\text{DTD}_{\text{HDDM\text{-}W}}$ & Baseline & $\text{DTD}_{\text{PUDD\text{-}1}}$ & Baseline & $\text{DTD}_{\text{PUDD\text{-}3}}$ & Baseline & $\text{DTD}_{\text{PUDD\text{-}5}}$ \\ \midrule
\multirow{2}{*}{Airline} & \small continual & 60.55$\pm$1.65 & \textbf{64.59$\pm$0.39} & \textbf{63.31$\pm$0.52} & 62.49$\pm$0.31 & \textbf{63.21$\pm$0.36} & 62.60$\pm$0.29 & \textbf{63.35$\pm$0.50} & 62.53$\pm$0.34 \\
 & \small sporadic & 61.81$\pm$0.42 & \textbf{62.28$\pm$0.39} & \textbf{60.90$\pm$0.08} & 60.73$\pm$0.91 & \textbf{60.16$\pm$0.57} & 59.99$\pm$1.21 & \textbf{60.19$\pm$0.86} & 59.77$\pm$0.89 \\
\multirow{2}{*}{Elec2} & \small continual & 73.33$\pm$1.36 & \textbf{75.30$\pm$0.34} & 74.92$\pm$0.18 & \textbf{76.57$\pm$0.50} & 74.93$\pm$0.90 & \textbf{76.64$\pm$0.29} & 74.92$\pm$0.41 & \textbf{76.58$\pm$0.34} \\
 & \small sporadic & 72.82$\pm$1.66 & \textbf{73.05$\pm$0.52} & 69.35$\pm$0.19 & \textbf{71.47$\pm$0.76} & 68.98$\pm$0.67 & \textbf{71.68$\pm$0.69} & 68.68$\pm$0.24 & \textbf{71.54$\pm$0.80} \\
\multirow{2}{*}{PS} & \small continual & 70.20$\pm$3.05 & \textbf{72.19$\pm$0.16} & \textbf{72.25$\pm$0.74} & 72.12$\pm$0.17 & \textbf{72.23$\pm$0.57} & 72.14$\pm$0.19 & \textbf{72.24$\pm$0.10} & 72.21$\pm$0.19 \\
 & \small sporadic & 68.58$\pm$3.42 & \textbf{71.89$\pm$0.33} & \textbf{71.47$\pm$0.31} & 71.02$\pm$0.79 & 70.37$\pm$0.92 & \textbf{70.85$\pm$0.55} & 70.20$\pm$0.72 & \textbf{70.69$\pm$0.70} \\
\multirow{2}{*}{SEA0} & \small continual & 97.96$\pm$0.21 & \textbf{98.77$\pm$0.11} & 97.94$\pm$0.89 & \textbf{98.34$\pm$0.17} & 98.04$\pm$0.15 & \textbf{98.35$\pm$0.17} & 98.23$\pm$0.03 & \textbf{98.35$\pm$0.18} \\
 & \small sporadic & 92.21$\pm$1.07 & \textbf{93.91$\pm$1.76} & 94.89$\pm$0.37 & \textbf{97.90$\pm$0.16} & 95.99$\pm$0.79 & \textbf{97.87$\pm$0.31} & 96.29$\pm$0.22 & \textbf{97.92$\pm$0.17} \\
\multirow{2}{*}{SEA10} & \small continual & 88.05$\pm$0.43 & \textbf{89.11$\pm$0.10} & 87.86$\pm$0.85 & \textbf{88.02$\pm$0.24} & 87.80$\pm$0.15 & \textbf{88.02$\pm$0.24} & 87.80$\pm$0.61 & \textbf{88.04$\pm$0.29} \\
 & \small sporadic & 85.30$\pm$0.55 & \textbf{86.39$\pm$0.51} & 85.91$\pm$0.26 & \textbf{86.93$\pm$0.39} & 86.02$\pm$0.20 & \textbf{87.01$\pm$0.31} & 86.24$\pm$0.07 & \textbf{86.89$\pm$0.37} \\
\multirow{2}{*}{SEA20} & \small continual & 77.50$\pm$1.58 & \textbf{79.14$\pm$0.11} & \textbf{77.50$\pm$0.57} & 77.45$\pm$0.38 & 77.33$\pm$0.06 & \textbf{77.38$\pm$0.31} & \textbf{77.38$\pm$0.30} & 77.36$\pm$0.34 \\
 & \small sporadic & \textbf{76.85$\pm$0.34} & 76.73$\pm$0.37 & 76.00$\pm$0.79 & \textbf{76.72$\pm$0.30} & 76.35$\pm$0.17 & \textbf{76.64$\pm$0.40} & 76.38$\pm$0.28 & \textbf{76.49$\pm$0.38} \\
\multirow{2}{*}{Sine} & \small continual & 87.56$\pm$7.46 & \textbf{94.19$\pm$0.74} & 86.19$\pm$0.88 & \textbf{93.02$\pm$1.07} & 85.12$\pm$0.82 & \textbf{93.26$\pm$0.91} & 82.51$\pm$0.48 & \textbf{93.23$\pm$0.92} \\
 & \small sporadic & 87.07$\pm$2.44 & \textbf{88.57$\pm$2.70} & 83.39$\pm$0.34 & \textbf{90.63$\pm$1.79} & 84.97$\pm$0.17 & \textbf{91.36$\pm$0.97} & 85.09$\pm$0.15 & \textbf{91.33$\pm$0.96} \\
\multirow{2}{*}{Mixed} & \small continual & 80.95$\pm$8.86 & \textbf{89.98$\pm$0.02} & 77.39$\pm$0.92 & \textbf{88.79$\pm$0.13} & 80.05$\pm$0.70 & \textbf{88.79$\pm$0.12} & 82.81$\pm$0.02 & \textbf{88.79$\pm$0.12} \\
 & \small sporadic & 85.21$\pm$0.48 & \textbf{86.92$\pm$0.16} & 82.65$\pm$0.04 & \textbf{86.34$\pm$1.64} & 84.65$\pm$0.02 & \textbf{86.92$\pm$0.18} & 84.90$\pm$0.39 & \textbf{86.92$\pm$0.18} \\
\bottomrule
\end{tabular}

\caption{Performance comparison with classic drift detector and PUDD using the \textbf{DNN} classifier. We compare each baseline against our proposed method $\text{DTD}_{\text{Baseline}}$. The results are presented as mean accuracy (\%) $\pm$ standard deviation (multiplied by 100 for space efficiency). The best performance in each pair is highlighted in \textbf{bold}. PS is short for powersupply.}
\label{tab:dnn_results}
\end{table*}

\section{Experiment}

\begin{table}[!t]
\centering
\small % Use 9-point font.
\setlength{\tabcolsep}{1mm} % Reduce column separation.

\begin{tabular}{@{}lcccccccl@{}}
\toprule
\textbf{Method} & \textbf{Airline} & \textbf{Elec2} & \textbf{Mixed} & \textbf{PS} & \textbf{SEA0} & \textbf{SEA10} & \textbf{SEA20} \\
\midrule
AMF             & 38.56          & 66.24          & 49.49          & 69.63          & 93.67          & 83.70          & 73.41          \\
IWE             & 38.02          & 68.90          & 49.47          & 64.10          & 93.14          & 84.73          & 74.33          \\
NS              & 67.91          & 76.42          & 81.09          & 72.39          & 93.54          & 84.39          & 76.00          \\
ADLTER          & \textbf{70.00} & 76.10          & 87.63          & 72.48          & 93.40          & 85.89          & 76.48          \\
MCD-DD          & 63.65          & 69.81          & 86.68          & 71.66          & 97.66          & 87.22          & 77.25          \\
PUDD-1            & 63.78          & 77.28          & 89.51          & 72.68          & 98.47          & 87.72          & 76.93          \\
PUDD-3            & 64.62          & 76.77          & 89.47          & \textbf{72.79} & 98.44          & 87.67          & 77.22          \\
PUDD-5            & 64.45          & 76.92          & 89.37          & 72.74          & 98.49          & 87.74          & 77.32          \\
\midrule
$\text{DTD}_\text{PUDD-1}$       & 65.59          & \textbf{77.30} & 89.92          & 72.33          & 98.70          & \textbf{88.88} & \textbf{78.86} \\
$\text{DTD}_\text{PUDD-3}$       & 65.59          & 77.16          & 89.92          & 72.34          & \textbf{98.71} & \textbf{88.88} & 78.85          \\
$\text{DTD}_\text{PUDD-5}$       & 65.62          & \textbf{77.30} & \textbf{89.98} & 72.33          & \textbf{98.71} & \textbf{88.88} & 78.84          \\
\bottomrule
\end{tabular}

\caption{Test accuracy (\%) comparison of DTD VS SOTA methods on various real-world and synthetic datasets. The best-performing method is in bold. PS is short for powersupply. Result on Sine dataset is provided in Appendix.}
\label{tab:accuracy_comparison_updated}
\end{table}

\subsection{Datasets and Baselines}
Our experimental evaluation is conducted on a diverse set of datasets, comprising 3 real-world datasets (airline \cite{airline}, elec2 \cite{elec}, powersupply \cite{powersupply}) and 4 synthetic datasets (sine \cite{ddm}, mixed \cite{ddm}, CIFAR-10-CD \cite{aaai}, sea variants \cite{moa}). Dataset details are provided in Appendix.

We establish comprehensive comparisons against 11 baseline methods: 5 classic concept drift detectors and 6 state-of-the-art (SOTA) approaches.
The classic detectors include DDM \cite{ddm}, HDDM-A \cite{HDDM}, HDDM-W \cite{HDDM}, KSWIN \cite{kswin}, and PH \cite{ph}.
The SOTA methods are MCDD \cite{mcdd}, AMF \cite{amf}, IWE \cite{iwe}, NS \cite{ns}, ADLTER \cite{adlter}, and PUDD \cite{aaai}. The threshold of these methods are set according to their original papers. The original PUDD paper proposes three options for the initial threshold. To distinguish between them, we denote a PUDD detector with an initial threshold of $10^{-x}$ as PUDD-$x$. Specific details of all baselines are described in Appendix due to page limit.

For baseline methods that are inherently classifier-agnostic, namely DDM, HDDM-A, HDDM-W, KSWIN, PH, and PUDD, we evaluate our proposed method using different base classifiers to ensure a comprehensive comparison. Specifically, our method is configured with a Gaussian Naive Bayes classifier (GNB) , a Very Fast Decision Tree (VFDT) \cite{vfdt}, and a Deep Neural Network (DNN) as its base learners. We also test all methods with two different training scenario for comprehensive evaluation. In the continual scenario, the classifier learns at every time step. In the sporadic scenario, the classifier is trained only upon a drift alarm. 

For the remaining SOTA methods (MCDD, AMF, IWE, NS, ADLTER), which primarily raise adaption without adaption, we compare them directly against our proposed ensemble method. Our method is presented as an ensemble version of PUDD, further enhanced with our novel DTD mechanism, considering that these baselines are also ensemble-based approaches. More implementation details are provided in Appendix due to page limit.

\subsection{Comparison with Baselines and Ablation Studies}

We conduct a comprehensive set of experiments to validate our claims and evaluate the performance of our proposed Dynamic Threshold Determination (DTD) Algorithm. Our evaluation is threefold: first, we apply DTD to a wide range of established drift detectors to demonstrate its general compatibility and effectiveness. Second, we compare a DTD-enhanced detector against state-of-the-art (SOTA) concept drift handling methods. Third, we test its applicability on complex image data streams. The results are summarized in Table \ref{tab:gnb_results}, \ref{tab:dnn_results}, \ref{tab:accuracy_comparison_updated}, and Figure \ref{fig:cifar10_bar_chart}. The result on VFTD classifier is provided in Appendix due to page limit. From these results, we draw several key observations.

\begin{figure}[t]
  \centering
  \includegraphics[width=\linewidth]{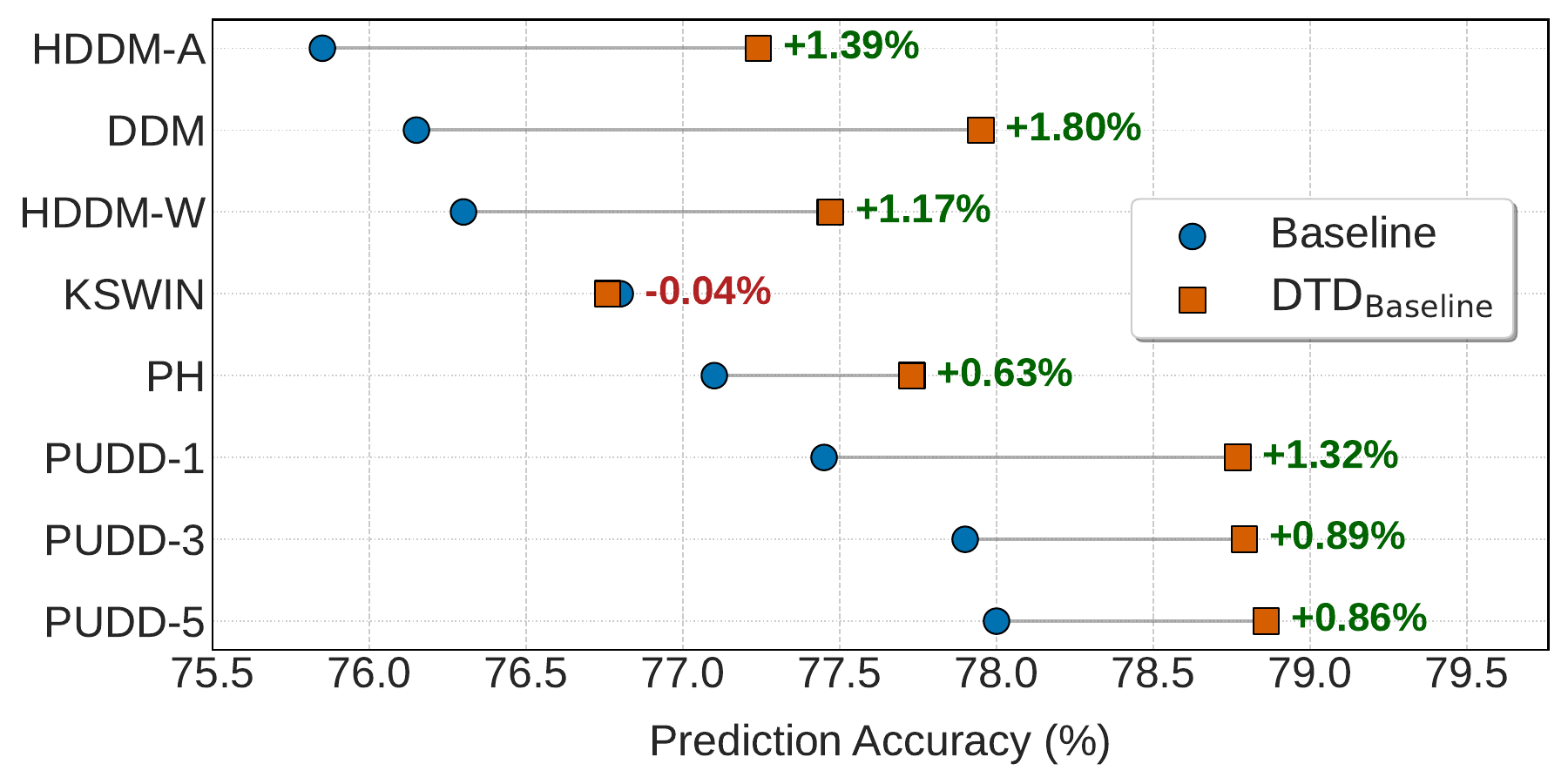}
  \caption{Comparison of accuracy on CIFAR10-CD dataset.}
  \label{fig:cifar10_bar_chart}
\end{figure}

\begin{figure}[t]
    \centering
    \includegraphics[width=\linewidth]{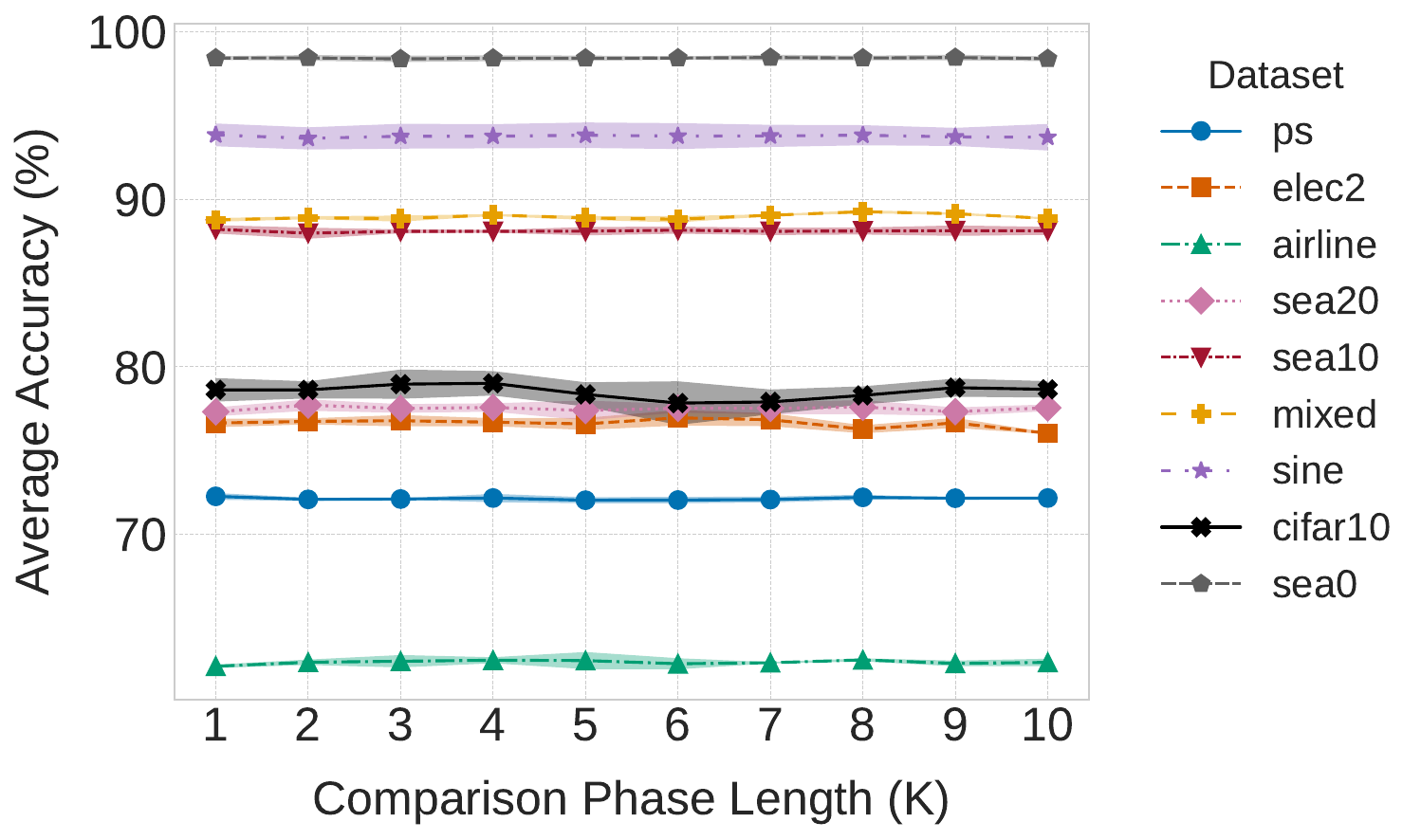}
    \caption{Ablation study of the $\text{DTD}_\text{PUDD}$ algorithm's hyper-parameter $K$, which indicates the length of comparison phase. Lines indicate the mean accuracy for each dataset, while shaded regions show the standard deviation calculated from multiple trials. PS is short for the powersupply dataset.}
    \label{fig:ablation_k}
\end{figure}

\paragraph{Observation 1: DTD generally improves the performance of existing drift detectors.}
Experimental results demonstrate that DTD significantly enhances the performance of existing drift detectors. As shown in Tables \ref{tab:gnb_results} and \ref{tab:dnn_results}, detectors equipped with DTD achieve higher predictive accuracy in the vast majority of scenarios. For instance, on the Sine dataset with a DNN classifier (Table~\ref{tab:dnn_results}), DTD boosts KSWIN's accuracy from 77.74\% to 94.14\%. While several exceptions exist where a baseline detector performs slightly better (e.g., PUDD-1 on Airline), this typically occurs when a candidate model's performance is high by chance during the brief comparison phase, leading to a suboptimal adjustment. Nevertheless, the evidence strongly indicates that our method is more robust and effective on average than relying on a fixed threshold.

\paragraph{Observation 2: The benefits of dynamic thresholding are more significant in complex scenarios.}
A closer analysis of the results reveals a notable trend regarding the performance of DTD. The gains are often more significant when it is paired with complex models or applied to more challenging datasets. While DTD provides benefits with simpler models like GNB and Hoeffding Trees, the improvements are particularly pronounced with the DNN classifier, as shown in Table \ref{tab:dnn_results}. On the Mixed and Sine datasets, DTD provides a substantial accuracy improvement to nearly every detector. This suggests that the limitations of a static threshold become more evident as data and model complexity increase, making an adaptive approach like DTD more critical. We further validate this on the challenging CIFAR10-CD image dataset (Figure \ref{fig:cifar10_bar_chart}). DTD yields consistent accuracy gains for almost all detectors, e.g., it improves DDM by +1.80\% and PUDD-1 by +1.32\%, with only a statistically insignificant drop of 0.04\% for KSWIN. This reinforces our argument that a dynamic, performance-aware thresholding mechanism is essential for handling complex data streams.

\begin{figure}[t]
  \centering
  \includegraphics[width=\linewidth]{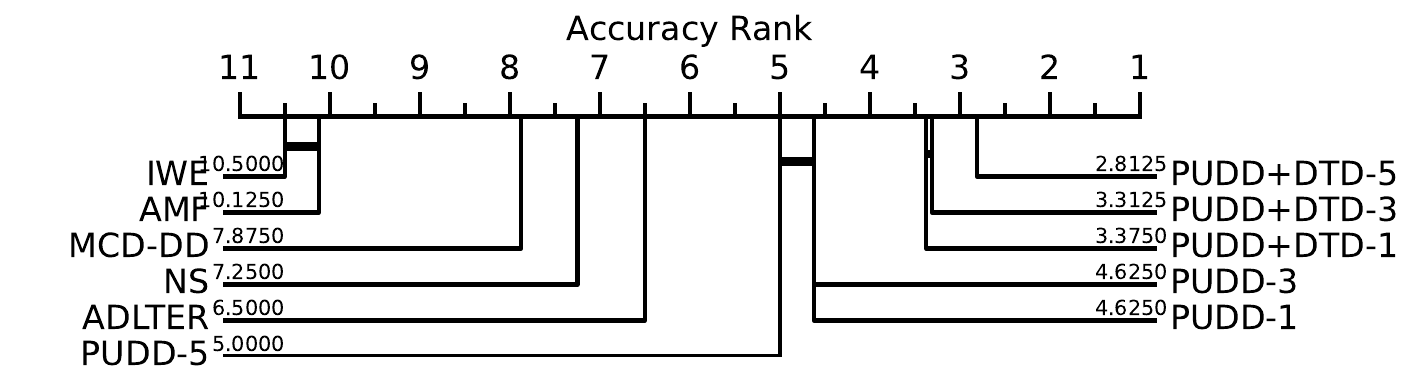}
  \caption{The critical difference diagram shows statistically significant superiority of $\text{DTD}_\text{PUDD}$ over SOTA methods.}
  \label{fig:cd-diagram}
\end{figure}

\paragraph{Observation 3: DTD is highly competitive with state-of-the-art methods.}
To evaluate its competitiveness, we compare our best configuration, $\text{DTD}_\text{PUDD}$, against several SOTA methods in Table~\ref{tab:accuracy_comparison_updated}. Our approach achieves the highest accuracy on all datasets, and consistently outperforms all rivals on the most cases, underscoring its robustness. Note the result on Sine dataset is provided in Appendix due to page limit. However, no single method dominates across all scenarios. For instance, ADLTER performs best on Airline, while a baseline PUDD-3 excels on PowerSupply and Sine. This observation reinforces our central thesis: no single configuration is universally optimal.  Nevertheless, the Wilcoxon-Holm analysis in Figure~\ref{fig:cd-diagram} shows $\text{DTD}_\text{PUDD}$ outperforms all SOTA competitors. 

\paragraph{Observation 4: DTD is robust to its comparison phase duration, $K$.}
To assess DTD's sensitivity to its main hyperparameter $K$, we performed a dedicated ablation study. As DTD is a threshold adaptation algorithm, it must be paired with a base detector to monitor for drift. For this analysis, we therefore selected the combination where DTD proved most effective: the PUDD detector with an initial threshold of $10^{-3}$ and a DNN classifier. The only exception was the \text{CIFAR10-CD} dataset, for which a ResNet-18 classifier was used. This configuration was evaluated across 9 diverse datasets. The results in Figure~\ref{fig:ablation_k} reveal remarkable stability for $K$ values in the range of $[1, 10]$. On most datasets, like \text{powersupply}, accuracy remains nearly constant around 72.1\%, showing the choice of $K$ has a negligible impact. Even on complex datasets such as \text{mixed} and \text{CIFAR10-CD}, performance variation is minimal, with accuracy on \text{mixed} fluctuating only between 88.7\% and 89.2\%. This result shows our method is insensitive to the choice of $K$. A small default value (e.g., $K=3$) thus provides a reliable and efficient configuration. This experiment validates the stability of our threshold adaptation mechanism.

\section{Conclusion}

This paper argues that conventional static thresholds for concept drift detection are suboptimal because they fail to maximize overall model performance. We theoretically prove that no single threshold is universally optimal and that dynamic strategies are inherently superior.
We demonstrate this by constructing a superior dynamic strategy from a sequence of locally optimal thresholds and proving that no single, static threshold can match its overall performance.

To address this, we propose a {Dynamic Threshold Determination Algorithm (DTD)}, which dynamically adjusts the detection threshold by empirically evaluating the performance of different online adaptation strategies. Our extensive experiments confirm that DTD consistently improves a wide range of existing detectors. Our DTD-enhanced detectors are highly competitive with SOTA methods.

Our future work will focus on two main objectives.
First, we will try to include the threshold threshold in a loss function to build an end-to-end framework for dynamic threshold determination.
Second, we plan to extend our algorithm to determine when to fine-tune large pre-trained language models. This would bring a cost-effective strategy that preserves model performance by avoiding unnecessary retraining.

\section{Acknowledges}
This work was supported by the Australian Research Council through the Laureate Fellow Project under Grant FL190100149.

\bibliography{bib}

\section{Appendix}

\begin{algorithm}[h]
\caption{CreateCandidates}
\begin{algorithmic}[1]
\STATE \textbf{Input:} $M, M', C_{curr}, C_{prev}, a, S_{curr}, S_{prev}, \psi$ % Changed \psi to \psi to avoid conflict
\STATE $\Psi = \{\texttt{PM}: \emptyset, \texttt{RDM}:\psi, \texttt{EDM}: \emptyset\}$ % Renamed \Psi to \Psi
\STATE Set $\Psi[\texttt{EDM}]$ as a new detector with threshold as $S_{prev}$.
\STATE Set $\Psi[\texttt{PM}]$ as a new detector with threshold as $S_{curr}+\eta$
\STATE $\mathcal{M} = \{\texttt{EDM}: M', \texttt{RDM}:\text{copy}(M),\texttt{PM}:\text{copy}(M)\}$ % Renamed \mathcal{M}
\STATE $\mathcal{M}[\texttt{RDM}] = \text{Adaption}(\mathcal{M}[\texttt{RDM}], C_{curr})$
\STATE $\mathcal{M}[\texttt{EDM}] = \text{Adaption}(\mathcal{M}[\texttt{EDM}], C_{prev})$
\STATE $\text{Train}(\mathcal{M}[\texttt{PM}], C_{curr})$
\STATE $\Pi_{cand} = \{\texttt{PM}: [a], \texttt{RDM}: [a], \texttt{EDM}: []\}$ % Renamed \Pi
\STATE $a', s = \text{Evaluate}(\mathcal{M}[\texttt{EDM}], C_{curr}, \Psi[\texttt{EDM}])$
\STATE Add $a'$ to $\Pi_{cand}[\texttt{EDM}]$
\IF{$s>S_{prev}$}
    \STATE $\mathcal{M}[\texttt{EDM}]=\text{Adaption}(\mathcal{M}[\texttt{EDM}], C_{curr})$
\ELSE
    \STATE $\text{Train}(\mathcal{M}[\texttt{EDM}], C_{curr})$ \# If continual training
\ENDIF
\STATE \textbf{return} $\mathcal{M}, \Pi_{cand}, \Psi$
\end{algorithmic}
\label{algo-2}
\end{algorithm}

\begin{algorithm}[h]
\caption{EvalCandidates}
\label{algo-3}
\begin{algorithmic}[1]
% In algorithmic, you usually list the steps.
% You can optionally add an input line to show the signature.
\STATE \textbf{Input:} $\mathcal{M}, C_{curr}, \Pi, \Psi$ % Representing the procedure signature
    \STATE $A_{step} = \{\}$
    \FOR{$(\texttt{name}, M) \in \mathcal{M}$}
        \STATE $a, s = \text{Evaluate}(\mathcal{M}[\texttt{name}], C_{curr}, \Psi[\texttt{name}])$
        \STATE $A_{step}[\texttt{name}] = a$; Add $a$ to $\Pi[\texttt{name}]$
        \IF {$s < \text{threshold  of } \Psi[\texttt{name}]$}
            \STATE $\mathcal{M}[\texttt{name}]=\text{Adaption}(\mathcal{M}[\texttt{name}], C_{curr})$
        \ELSE
            \STATE $\text{Train}(\mathcal{M}[\texttt{name}], C_{curr})$ \# If continual training
        \ENDIF
    \ENDFOR
    \STATE \textbf{return} $A_{step}$
\end{algorithmic}
\end{algorithm}

\subsection{Implementation details}

Our empirical validation is designed around four key analyses: comparisons against classic and state-of-the-art (SOTA) drift detectors, a benchmark on the CIFAR-10-CD dataset, and a concluding ablation study. All experiments were executed on an Ubuntu 18.04 server equipped with an NVIDIA A100 GPU and 200GB of RAM. To ensure statistical robustness, we report the average accuracy over 20 independent runs with distinct random seeds for every experiment.

\bigskip % Adds a vertical space for separation

\textbf{Classifiers and Architectures.} We employed three types of classifiers: a Gaussian Naive Bayes (GNB) from \texttt{scikit-learn} \cite{scikit-learn}, a Hoeffding Tree (VFDT) from the \texttt{River} package \cite{river}, and a Deep Neural Network (DNN) implemented in \texttt{PyTorch} \cite{pytorch}. The DNN, a Multi-Layer Perceptron (MLP), was configured with two primary architectures based on dataset complexity.
\begin{itemize}
    \item \textbf{Default Architecture:} For most datasets, the network consists of two hidden layers, each containing 64 neurons with ReLU activation. The input dimension is tailored to the specific dataset.
    \item \textbf{Airline Dataset Architecture:} For this more complex dataset, we utilized a deeper network with three hidden layers of 512, 256, and 64 neurons, respectively, all using ReLU activation, to handle an input dimension of 679.
\end{itemize}
In all DNNs, a final fully connected layer maps the hidden representation to the dataset-specific output dimension.

\bigskip % Adds a vertical space for separation

\textbf{Evaluation Protocols and Adaptation.} The procedures were specifically configured for different experimental tracks.

\textit{For the classic comparison and ablation study}, data is processed in chunks. The classifier's error rate on each chunk is fed to the drift detector. We assess performance under two schemes: an incremental \textit{test-then-train} approach and a \textit{train-once-until-alarm} setting called sporadic in this paper. The adaptation strategy depends on the classifier type: GNB and VFDT are retrained from scratch on the new data, while the DNN's final layer is reset. For its training, the DNN uses the Adam optimizer for 100 epochs with a learning rate of $1 \times 10^{-2}$.

\textit{For the state-of-the-art (SOTA) comparison}, we adopted an ensemble approach to ensure fairness against leading SOTA methods. Our model is an ensemble of 5 base DNNs, using soft voting for prediction and combined uncertainty for drift detection. The training protocol was adjusted accordingly: the Adam optimizer's learning rate was increased to $5 \times 10^{-2}$, and upon drift detection, the final layer of each DNN was fine-tuned for 20 epochs rather than being reset.

\textit{For the comparison on the CIFAR-10-CD image dataset}, we employed a ResNet-18 model as the classifier. Given the high learning difficulty of this benchmark, evaluation was conducted purely in an incremental fashion (\textit{test-then-train}). The model was optimized using Stochastic Gradient Descent (SGD) with a learning rate of $1 \times 10^{-2}$ and was trained for 5 epochs at each adaptation step.

We benchmark our method against a comprehensive suite of classic and state-of-the-art drift detectors. These include methods that monitor classifier performance, such as \textbf{DDM}~\cite{ddm}, which triggers an alarm when the error rate exceeds a threshold, and \textbf{EDDM}~\cite{EDDM}, which specializes in gradual changes by measuring the distance between errors. Another category relies on statistical tests within a data window. \textbf{KSWIN}~\cite{kswin} uses the Kolmogorov-Smirnov test, while \textbf{HDDM}~\cite{HDDM} offers two variants: \textbf{HDDM-A} for abrupt changes using moving averages and \textbf{HDDM-W} for gradual changes with weighted averages. \textbf{PH}~\cite{ph} enhances the Page Hinkley Test for more robust detection without manual tuning.

We also consider compare our method with SOTA models. \textbf{IWE}~\cite{iwe} incrementally re-weights historical classifiers in a variable-size window. For tree ensembles, \textbf{AMF}~\cite{amf} is an online Random Forest that prunes its trees, while \textbf{ADLTER}~\cite{adlter} and \textbf{NS}~\cite{ns} are designed for Gradient Boosting Decision Trees (GBDTs); the former adapts the number of iterations while the latter prunes weak learners. Finally, we include \textbf{MCDD}~\cite{mcdd}, a modern method that uses contrastive learning and concept discrepancy to identify drift in high-dimensional data streams.

Our evaluation is conducted on a diverse suite of benchmarks, comprising both real-world and synthetic datasets, to thoroughly assess performance under various drift conditions.

The real-world datasets include \textbf{Elec2}~\cite{elec}, from the Australian electricity market, which contains 45,000 instances partitioned into 45 chunks, with features representing electricity demand and a binary label indicating price direction. We also use the \textbf{Airline} dataset~\cite{airline}, consisting of 58,000 flight records across 58 chunks, where the task is to predict flight delays. Following the procedure in~\cite{driftsurf}, this dataset is one-hot encoded, expanding its feature dimension to 679. The \textbf{PowerSupply} dataset~\cite{powersupply} provides 29,000 hourly power records over 29 chunks, exhibiting drift from seasonal and weekly patterns. Lastly, we use \textbf{CIFAR-10-CD}~\cite{aaai}, a modification of the CIFAR-10 image dataset designed to simulate concept drift. It contains 50,000 images in 100 chunks, with labels that evolve according to a Markov process to reflect changing user interests.

\begin{table*}[!t]
\centering
\small
\setlength{\tabcolsep}{1mm} % 调整列间距

% The two parts of the table are placed side-by-side
% Part 1
\begin{tabular}{ @{} ll cc cc cc cc @{} }
\toprule
\multirow{2}{*}{\textbf{Dataset}} & \multirow{2}{*}{\textbf{\begin{tabular}[c]{@{}c@{}}Training\end{tabular}}} & \multicolumn{2}{c}{\textbf{KSWIN}} & \multicolumn{2}{c}{\textbf{DDM}} & \multicolumn{2}{c}{\textbf{PH}} & \multicolumn{2}{c}{\textbf{HDDM-A}} \\
\cmidrule(lr){3-4} \cmidrule(lr){5-6} \cmidrule(lr){7-8} \cmidrule(lr){9-10}
 &  & Baseline & $\text{DTD}_{\text{KSWIN}}$ & Baseline & $\text{DTD}_{\text{DDM}}$ & Baseline & $\text{DTD}_{\text{PH}}$ & Baseline & $\text{DTD}_{\text{HDDM\text{-}A}}$ \\ \midrule
\multirow{2}{*}{Airline} & \small continual & 61.21$\pm$0.53 & \textbf{61.76$\pm$0.40} & 60.16$\pm$0.00 & \textbf{60.34$\pm$0.00} & 60.95$\pm$0.00 & \textbf{61.95$\pm$0.00} & 60.95$\pm$0.00 & \textbf{61.35$\pm$0.00} \\
 & \small sporadic & \textbf{60.88$\pm$2.14} & 60.87$\pm$5.32 & 59.28$\pm$0.00 & \textbf{62.27$\pm$0.00} & 60.97$\pm$0.00 & \textbf{62.07$\pm$0.00} & 60.29$\pm$0.00 & \textbf{62.11$\pm$0.00} \\
\multirow{2}{*}{Elec2} & \small continual & 74.14$\pm$0.09 & \textbf{74.55$\pm$0.48} & \textbf{74.82$\pm$0.00} & 72.89$\pm$0.00 & 73.70$\pm$0.00 & \textbf{74.39$\pm$0.00} & 73.90$\pm$0.00 & \textbf{74.00$\pm$0.00} \\
 & \small sporadic & 74.10$\pm$0.08 & \textbf{74.25$\pm$0.83} & \textbf{74.75$\pm$0.00} & 74.25$\pm$0.00 & \textbf{73.99$\pm$0.00} & 72.28$\pm$0.00 & 73.83$\pm$0.00 & \textbf{74.92$\pm$0.00} \\
\multirow{2}{*}{PS} & \small continual & \textbf{71.32$\pm$0.23} & 71.18$\pm$0.24 & 70.68$\pm$0.00 & \textbf{72.05$\pm$0.00} & 70.88$\pm$0.00 & \textbf{71.18$\pm$0.00} & 70.84$\pm$0.00 & \textbf{71.91$\pm$0.00} \\
 & \small sporadic & 68.69$\pm$1.82 & \textbf{70.87$\pm$0.71} & 67.53$\pm$0.00 & \textbf{70.74$\pm$0.00} & 68.67$\pm$0.00 & \textbf{70.12$\pm$0.00} & 71.24$\pm$0.00 & \textbf{72.10$\pm$0.00} \\
\multirow{2}{*}{SEA0} & \small continual & 93.03$\pm$0.80 & \textbf{93.59$\pm$0.64} & 94.90$\pm$0.49 & \textbf{95.45$\pm$0.33} & 94.87$\pm$0.29 & \textbf{95.36$\pm$0.26} & 95.33$\pm$0.25 & \textbf{95.44$\pm$0.27} \\
 & \small sporadic & 89.32$\pm$2.94 & \textbf{90.18$\pm$2.95} & 93.44$\pm$0.87 & \textbf{94.26$\pm$0.67} & 93.95$\pm$0.59 & \textbf{94.61$\pm$0.24} & 93.84$\pm$0.57 & \textbf{94.51$\pm$0.29} \\
\multirow{2}{*}{SEA10} & \small continual & 84.49$\pm$0.67 & \textbf{85.29$\pm$0.75} & 85.06$\pm$0.62 & \textbf{86.19$\pm$0.70} & 86.03$\pm$0.24 & \textbf{86.75$\pm$0.27} & 85.86$\pm$0.37 & \textbf{86.42$\pm$0.41} \\
 & \small sporadic & 82.98$\pm$1.55 & \textbf{84.57$\pm$1.32} & 84.71$\pm$1.20 & \textbf{86.38$\pm$0.68} & 86.54$\pm$0.30 & \textbf{87.10$\pm$0.29} & 86.11$\pm$0.43 & \textbf{86.75$\pm$0.39} \\
\multirow{2}{*}{SEA20} & \small continual & 76.02$\pm$0.51 & \textbf{76.94$\pm$0.62} & 75.92$\pm$0.52 & \textbf{76.65$\pm$0.47} & 76.66$\pm$0.18 & \textbf{77.30$\pm$0.22} & 76.53$\pm$0.28 & \textbf{77.12$\pm$0.31} \\
 & \small sporadic & 74.75$\pm$0.96 & \textbf{76.90$\pm$0.65} & 74.97$\pm$1.21 & \textbf{76.69$\pm$0.76} & 76.83$\pm$0.52 & \textbf{77.59$\pm$0.28} & 76.33$\pm$0.58 & \textbf{77.15$\pm$0.58} \\
\multirow{2}{*}{Sine} & \small continual & 86.13$\pm$0.87 & \textbf{86.84$\pm$0.79} & 86.95$\pm$1.21 & \textbf{87.99$\pm$1.26} & 87.17$\pm$1.37 & \textbf{88.16$\pm$1.09} & 87.66$\pm$1.34 & \textbf{88.00$\pm$1.19} \\
 & \small sporadic & 82.21$\pm$2.24 & \textbf{83.75$\pm$1.90} & 82.74$\pm$4.63 & \textbf{85.30$\pm$2.34} & 85.43$\pm$1.81 & \textbf{86.50$\pm$1.12} & 85.28$\pm$1.84 & \textbf{86.16$\pm$1.61} \\
\multirow{2}{*}{Mixed} & \small continual & 84.41$\pm$0.11 & \textbf{84.73$\pm$0.11} & 84.13$\pm$0.19 & \textbf{84.57$\pm$0.34} & 83.72$\pm$0.41 & \textbf{84.72$\pm$0.17} & 84.41$\pm$0.11 & \textbf{84.64$\pm$0.46} \\
 & \small sporadic & 83.50$\pm$0.28 & \textbf{83.92$\pm$0.20} & 82.67$\pm$2.21 & \textbf{83.82$\pm$0.60} & 83.50$\pm$0.28 & \textbf{83.92$\pm$0.20} & 83.50$\pm$0.28 & \textbf{83.92$\pm$0.20} \\
\bottomrule
\end{tabular}
\quad % 两个表格之间的间隔
% Part 2
\begin{tabular}{ @{} ll cc cc cc cc @{} }
\toprule
\multirow{2}{*}{\textbf{Dataset}} & \multirow{2}{*}{\textbf{\begin{tabular}[c]{@{}c@{}}Training\end{tabular}}} & \multicolumn{2}{c}{\textbf{HDDM-W}} & \multicolumn{2}{c}{\textbf{PUDD-1}} & \multicolumn{2}{c}{\textbf{PUDD-3}} & \multicolumn{2}{c}{\textbf{PUDD-5}} \\
\cmidrule(lr){3-4} \cmidrule(lr){5-6} \cmidrule(lr){7-8} \cmidrule(lr){9-10}
 &  & Baseline & $\text{DTD}_{\text{HDDM\text{-}W}}$ & Baseline & $\text{DTD}_{\text{PUDD\text{-}1}}$ & Baseline & $\text{DTD}_{\text{PUDD\text{-}3}}$ & Baseline & $\text{DTD}_{\text{PUDD\text{-}5}}$ \\ \midrule
\multirow{2}{*}{Airline} & \small continual & 61.11$\pm$0.00 & \textbf{61.55$\pm$0.00} & \textbf{61.38$\pm$0.82} & 61.17$\pm$0.15 & \textbf{61.57$\pm$0.06} & 61.19$\pm$0.16 & \textbf{61.57$\pm$0.01} & 61.19$\pm$0.16 \\
 & \small sporadic & 61.92$\pm$0.00 & \textbf{62.20$\pm$0.00} & 61.16$\pm$0.84 & \textbf{61.43$\pm$0.00} & \textbf{59.90$\pm$0.12} & 59.11$\pm$0.00 & 57.04$\pm$0.49 & \textbf{57.56$\pm$0.00} \\
\multirow{2}{*}{Elec2} & \small continual & 73.80$\pm$0.00 & \textbf{75.56$\pm$0.00} & 73.86$\pm$0.59 & \textbf{74.11$\pm$0.23} & 73.84$\pm$0.90 & \textbf{74.11$\pm$0.23} & 73.64$\pm$0.49 & \textbf{74.11$\pm$0.23} \\
 & \small sporadic & 73.73$\pm$0.00 & \textbf{73.94$\pm$0.00} & 69.79$\pm$0.45 & \textbf{72.95$\pm$0.01} & 69.79$\pm$0.47 & \textbf{72.95$\pm$0.01} & 71.83$\pm$0.67 & \textbf{72.95$\pm$0.01} \\
\multirow{2}{*}{PS} & \small continual & 70.99$\pm$0.00 & \textbf{72.27$\pm$0.00} & \textbf{71.77$\pm$0.06} & 71.66$\pm$0.17 & \textbf{71.79$\pm$0.77} & 71.66$\pm$0.17 & \textbf{71.79$\pm$0.21} & 71.66$\pm$0.17 \\
 & \small sporadic & 69.54$\pm$0.00 & \textbf{70.04$\pm$0.00} & \textbf{71.13$\pm$0.68} & 70.64$\pm$0.00 & \textbf{71.20$\pm$0.86} & 70.64$\pm$0.00 & 70.40$\pm$0.92 & \textbf{70.51$\pm$0.00} \\
\multirow{2}{*}{SEA0} & \small continual & 93.27$\pm$0.82 & \textbf{93.92$\pm$0.72} & 95.13$\pm$0.30 & \textbf{95.51$\pm$0.24} & 95.21$\pm$0.29 & \textbf{95.51$\pm$0.25} & 95.24$\pm$0.11 & \textbf{95.51$\pm$0.25} \\
 & \small sporadic & 91.65$\pm$0.97 & \textbf{92.53$\pm$1.04} & 94.10$\pm$0.72 & \textbf{94.63$\pm$0.30} & 94.63$\pm$0.25 & \textbf{94.64$\pm$0.27} & 94.56$\pm$0.92 & \textbf{94.62$\pm$0.27} \\
\multirow{2}{*}{SEA10} & \small continual & 85.54$\pm$0.36 & \textbf{86.47$\pm$0.33} & 86.35$\pm$0.31 & \textbf{86.66$\pm$0.20} & 86.25$\pm$0.09 & \textbf{86.65$\pm$0.23} & 86.19$\pm$0.32 & \textbf{86.60$\pm$0.23} \\
 & \small sporadic & 85.31$\pm$0.50 & \textbf{86.68$\pm$0.45} & 86.30$\pm$0.53 & \textbf{87.22$\pm$0.26} & 86.57$\pm$0.63 & \textbf{87.11$\pm$0.29} & 86.19$\pm$0.24 & \textbf{87.00$\pm$0.26} \\
\multirow{2}{*}{SEA20} & \small continual & 76.71$\pm$0.23 & \textbf{77.51$\pm$0.18} & 77.02$\pm$0.69 & \textbf{77.28$\pm$0.28} & 76.90$\pm$0.65 & \textbf{77.23$\pm$0.23} & 76.83$\pm$0.90 & \textbf{77.21$\pm$0.23} \\
 & \small sporadic & 76.86$\pm$0.33 & \textbf{77.38$\pm$0.40} & 76.85$\pm$0.73 & \textbf{77.46$\pm$0.41} & 76.86$\pm$0.30 & \textbf{77.51$\pm$0.40} & 76.29$\pm$0.47 & \textbf{77.29$\pm$0.58} \\
\multirow{2}{*}{Sine} & \small continual & 87.57$\pm$1.28 & \textbf{88.08$\pm$0.97} & 87.33$\pm$0.61 & \textbf{88.41$\pm$1.39} & 87.42$\pm$0.42 & \textbf{88.50$\pm$1.41} & 87.63$\pm$0.31 & \textbf{88.51$\pm$1.42} \\
 & \small sporadic & 85.27$\pm$1.31 & \textbf{86.38$\pm$1.26} & 82.21$\pm$0.47 & \textbf{86.40$\pm$1.25} & 85.81$\pm$0.49 & \textbf{86.41$\pm$1.17} & 86.01$\pm$0.29 & \textbf{86.47$\pm$1.12} \\
\multirow{2}{*}{Mixed} & \small continual & 84.41$\pm$0.11 & \textbf{84.73$\pm$0.13} & \textbf{84.25$\pm$0.31} & 83.96$\pm$0.50 & 83.94$\pm$0.92 & \textbf{84.02$\pm$0.60} & 84.01$\pm$0.40 & \textbf{84.05$\pm$0.52} \\
 & \small sporadic & 83.50$\pm$0.28 & \textbf{83.92$\pm$0.20} & 82.13$\pm$0.11 & \textbf{82.78$\pm$2.18} & \textbf{84.04$\pm$0.03} & 83.93$\pm$0.20 & \textbf{84.16$\pm$0.35} & 83.92$\pm$0.20 \\
\bottomrule
\end{tabular}

\caption{Performance comparison with classic drift detector and PUDD using the \textbf{VFDT} classifier. We compare each baseline our proposed method $\text{DTD}_{\text{Baseline}}$. The results are presented as mean accuracy (\%) $\pm$ standard deviation (multiplied by 100 for space efficiency). The best performance in each pair is highlighted in \textbf{bold}. PS is short for powersupply.}
\label{tab:ht_results_formatted}
\end{table*}

\begin{table*}[!t]
\centering
\small % 使用9号字体

% 由于转置后列数较多，可能需要调整列间距或字体大小以适应页面
\setlength{\tabcolsep}{2.5pt} % 调整列间距

\begin{tabular}{@{}lccccccccccc@{}}
\toprule
\textbf{Dataset} & \textbf{AMF} & \textbf{IWE} & \textbf{NS} & \textbf{ADLTER} & \textbf{MCD-DD} & \textbf{PUDD-1} & \textbf{PUDD-3} & \textbf{PUDD-5} & \textbf{$\text{DTD}_\text{PUDD-1}$} & \textbf{$\text{DTD}_\text{PUDD-3}$} & \textbf{$\text{DTD}_\text{PUDD-5}$} \\
\midrule
Sine & 49.52 & 49.51 & 91.01 & 92.18 & 90.21 & \textbf{94.52} & 94.76 & 90.90 & 94.01 & 94.06 & 94.07 \\
\bottomrule
\end{tabular}

\caption{Test accuracy (\%) comparison of DTD VS SOTA methods on various real-world and synthetic datasets. The best-performing method is in bold. PS is short for powersupply. Result on Sine dataset is provided in Appendix.}
\label{tab:sota2}
\end{table*}

To analyze performance in controlled settings, we employ three synthetic datasets, each with 100,000 instances divided into 100 chunks where an abrupt drift is induced every tenth chunk. The \textbf{SEA} dataset~\cite{moa} introduces drift by altering the classification function's thresholds. The \textbf{SINE} dataset~\cite{ddm}, defined by two attributes, generates drift by changing the sine function used for labeling. Finally, the \textbf{Mixed} dataset~\cite{ddm}, containing boolean and numeric features, simulates drift by changing its underlying classification function.

\clearpage
\onecolumn

\subsection{Proof of Theorem 1}

\begin{proof}
We substantiate this claim by constructing three counterexamples, each corresponding to a distinct type of concept drift (sudden, gradual, and recurrent).

\paragraph{Case 1: Sudden Drift.}
Suppose the stream has total length \(T\) and a single, abrupt change from concept \(C_1\) to concept \(C_2\). Concretely, the first \(t_d\) samples (indices \(1\) through \(t_d\)) follow \(C_1\), and starting from sample \((t_d+1)\), the data follow \(C_2\). We employ a threshold-based drift detector (e.g., DDM~\cite{ddm}) with warning threshold~\(\theta_1\) and confirm threshold~\(\theta_2\). Denote by \(M_t\) the classifier at time \(t\). The average accuracy over the entire stream is
\[
A \;=\; \frac{1}{T}\sum_{i=1}^T \mathrm{Acc}\bigl(M_t(x_t),\,y_t\bigr).
\]

\emph{Perfect detection} means the detector confirms the drift exactly at the moment it occurs, i.e., right before sample \((t_d+1)\). Nonetheless, there is an unavoidable \emph{one-sample mismatch} at index \((t_d+1)\), where a model still trained on \(C_1\) encounters the first sample from \(C_2\). Immediately after that mismatch, an incremental-adaptation phase of length \(t_{\mathrm{incre}}\) begins with average accuracy \(A_{\mathrm{incre}}\). The model then attains a stable accuracy \(A_{\mathrm{stable}}\) on \(C_2\) for the remaining \(\bigl(T - t_d - 1 - t_{\mathrm{incre}}\bigr)\) samples. Let \(A_{C_1}\) be the average accuracy on \(C_1\), and let \(A_{\mathrm{dismatch}}\) be the accuracy of the outdated \(C_1\)-based model on that single mismatch sample. Then the overall performance under perfect detection is
\begin{equation}
\label{eq:perfA_sudden}
\begin{aligned}
A_P 
&= 
\frac{1}{T}
\Bigl[
  t_d \, A_{C_1}
  \;+\;
  1 \cdot A_{\mathrm{dismatch}}
  \;+\;
  t_{\mathrm{incre}} \, A_{\mathrm{incre}}
  \;+\;
  \bigl(T - t_d - 1 - t_{\mathrm{incre}}\bigr)\,A_{\mathrm{stable}}
\Bigr].
\end{aligned}
\end{equation}

By contrast, under \emph{delayed detection}, the detector postpones confirmation so that the model remains mismatched for \(t_w\) consecutive samples (from \((t_d+1)\) to \((t_d + t_w)\)). The model then retrains on \(C_2\), adapts incrementally for \(t_{\mathrm{incre}}'\) samples (with average accuracy \(A_{\mathrm{incre}}'\)), and finally converges to \(A_{\mathrm{stable}}\). Thus, the performance in this case is
\begin{equation}
\label{eq:delayedA_sudden}
\begin{aligned}
A_D 
&= 
\frac{1}{T}
\Bigl[
  t_d \, A_{C_1}
  \;+\;
  t_w \, A_{\mathrm{dismatch}}
  \;+\;
  t_{\mathrm{incre}}' \, A_{\mathrm{incre}}'
  \;+\;
  \bigl(T - t_d - t_w - t_{\mathrm{incre}}'\bigr)\,A_{\mathrm{stable}}
\Bigr].
\end{aligned}
\end{equation}
Subtracting \eqref{eq:perfA_sudden} from \eqref{eq:delayedA_sudden}, one finds a sufficient condition for \(A_D > A_P\):
\[
(t_w - 1)\,A_{\mathrm{dismatch}}
\;+\;
\Bigl(t_{\mathrm{incre}}' \, A_{\mathrm{incre}}' \;-\; t_{\mathrm{incre}} \, A_{\mathrm{incre}}\Bigr)
\;+\;
\Bigl(1 + t_{\mathrm{incre}} - t_w - t_{\mathrm{incre}}'\Bigr)\,A_{\mathrm{stable}}
\;>\;
0.
\]
Hence, even an exact drift detector may be suboptimal if early adaptation is costly. Allowing a controlled mismatch of \(t_w\) instances can, in some scenarios, produce higher overall accuracy by enabling more efficient retraining.

\paragraph{Case 2: Gradual Drift.}
Now consider a stream of total length \(T\) in which the transition from \(C_1\) to \(C_2\) occurs gradually over \(t_g\) consecutive samples. That is, the first \(t_d\) samples follow \(C_1\), and from sample \((t_d+1)\) through sample \((t_d + t_g)\), the distribution shifts incrementally at each time step, ultimately settling on \(C_2\) from \((t_d + t_g + 1)\) onward.

Under \emph{perfect detection}, every incremental distribution change within \([t_d+1,\,t_d + t_g]\) is identified \emph{immediately} upon its occurrence. Consequently, the model retrains continually throughout those \(t_g\) samples, leaving no stable window for incremental learning. Let \(A_{\mathrm{g}}\) be the average accuracy over this continual-retraining phase. Once the drift fully completes at \((t_d + t_g)\), the model finally runs an adaptation phase of length \(t_{\mathrm{incre}}\) with average accuracy \(A_{\mathrm{incre}}\), then converges to \(A_{\mathrm{stable}}\). The performance under \emph{perfect detection} is
\[
\begin{aligned}
A_P 
&= 
\frac{1}{T}
\Bigl[
  t_d \, A_{C_1}
  \;+\;
  t_g \, A_{\mathrm{g}}
  \;+\;
  t_{\mathrm{incre}} \, A_{\mathrm{incre}}
  \;+\;
  \bigl(T - t_d - t_g - t_{\mathrm{incre}}\bigr)\,A_{\mathrm{stable}}
\Bigr].
\end{aligned}
\]

Under \emph{delayed detection}, by contrast, a warning may be raised at \((t_d+1)\) but the drift is not confirmed until \(\bigl(t_d + t_g + t_w\bigr)\). Hence, from sample \((t_d+1)\) through \((t_d + t_g + t_w)\), the model remains trained on \(C_1\) at mismatch accuracy \(A_{\mathrm{dismatch}}\). Only then does it retrain on \(C_2\), undergo an incremental-adaptation phase of length \(t_{\mathrm{incre}}'\) (with accuracy \(A_{\mathrm{incre}}'\)), and finally converge to \(A_{\mathrm{stable}}\). Thus,
\[
\begin{aligned}
A_D 
&= 
\frac{1}{T}
\Bigl[
  t_d \, A_{C_1}
  \;+\;
  (t_g + t_w)\,A_{\mathrm{dismatch}}
  \;+\;
  t_{\mathrm{incre}}' \, A_{\mathrm{incre}}'
  \;+\;
  \bigl(T - t_d - t_g - t_w - t_{\mathrm{incre}}'\bigr)\,A_{\mathrm{stable}}
\Bigr].
\end{aligned}
\]
Subtracting \(A_P\) from \(A_D\) and requiring \(A_D > A_P\) yields
\[
(t_g + t_w)\,A_{\mathrm{dismatch}}
\;-\;
t_g\,A_{\mathrm{g}}
\;+\;
\Bigl(t_{\mathrm{incre}}'\,A_{\mathrm{incre}}' \;-\; t_{\mathrm{incre}}\,A_{\mathrm{incre}}\Bigr)
\;+\;
\bigl(t_{\mathrm{incre}} - t_w - t_{\mathrm{incre}}'\bigr)\,A_{\mathrm{stable}}
\;>\; 
0.
\]
This inequality reflects the trade-off between tolerating a mismatch over \((t_g + t_w)\) samples at \(A_{\mathrm{dismatch}}\) and avoiding perpetual retraining. If the mismatch penalty is offset by maintaining one cohesive adaptation phase, the delayed strategy can outperform perfect detection even for a slow, piecewise drift.

\paragraph{Case 3: Recurrent Drift.}
Assume a stream of length \(T\). Concept \(C_1\) governs the first \(t_d\) samples. At time \((t_d + 1)\), the concept briefly switches to \(C_2\) for exactly one sample, then reverts to \(C_1\) from \((t_d + 2)\) onward. There are thus two drift points: moving from \(C_1\) to \(C_2\) at \((t_d + 1)\) and back to \(C_1\) at \((t_d + 2)\).

Under \emph{perfect detection}, the algorithm confirms both drifts immediately, causing four phases. First, the model is stably trained on \(C_1\) for the initial \(t_d\) samples, with average accuracy \(A_{C_1}\). Second, at sample \((t_d + 1)\), the classifier (still on \(C_1\)) encounters a one-sample mismatch against \(C_2\), yielding accuracy \(A_{\mathrm{dismatch}}\). Third, the concept shifts back to \(C_1\) at \((t_d + 2)\), but the model has just been adapted (or was in the process of adapting) to \(C_2\). It thus incurs one mismatched sample at \((t_d + 2)\), followed by an incremental-adaptation period of length \(t_{\mathrm{incre},1}\). Let the average accuracy over this mismatch-plus-adaptation block be split into two parts: the single mismatch sample at accuracy \(A_{\mathrm{mismatch2}}\) and the incremental-learning stage at accuracy \(A_{\mathrm{incre},1}\). Finally, once adaptation finishes, the classifier is again stable on \(C_1\) for the remaining \(\bigl(T - t_d - 2 - t_{\mathrm{incre},1}\bigr)\) samples, achieving accuracy \(A_{\mathrm{stable},1}\). Summing these four phases, the overall accuracy under perfect detection, \(A_P\), can be written as
\[
A_P
=
\frac{1}{T}\Bigl[
  t_d\,A_{C_1}
  \;+\;
  1 \cdot A_{\mathrm{dismatch}}
  \;+\;
  1 \cdot A_{\mathrm{mismatch2}}
  \;+\;
  t_{\mathrm{incre},1}\,A_{\mathrm{incre},1}
  \;+\;
  \bigl(T - t_d - 2 - t_{\mathrm{incre},1}\bigr)\,A_{\mathrm{stable},1}
\Bigr].
\]
Observe that the model invests in two distinct retraining episodes (one for the single sample of \(C_2\), then another immediately to revert to \(C_1\)).

In a \emph{missed detection} scenario, the brief appearance of \(C_2\) at \((t_d + 1)\) is never confirmed. The stream thus divides into three phases: stable on \(C_1\) for \(t_d\) samples, a single mismatch sample at \((t_d + 1)\) with accuracy \(A_{\mathrm{dismatch}}\), and a return to stable \(C_1\) from \((t_d + 2)\) onward with no retraining needed. Hence,
\[
A_M
=
\frac{1}{T}\Bigl[
  t_d\,A_{C_1}
  \;+\;
  1 \cdot A_{\mathrm{dismatch}}
  \;+\;
  \bigl(T - (t_d + 1)\bigr)\,A_{\text{stable,1}}
\Bigr].
\]
There is only one mismatch cost and no adaptation overhead. A straightforward comparison often shows that \(A_M\) can exceed \(A_P\), because the penalty from two retraining operations in the perfect-detection case outweighs the cost of ignoring a single-sample drift. This illustrates that detecting \emph{every} drift event may reduce overall accuracy when some drifts are too brief to warrant adaptation, aligning with the broader conclusion that perfect detection need not guarantee optimal performance in nonstationary learning.
\end{proof}

\subsection{Proof of Theorem 2}
\begin{proof}
Suppose, for contradiction, that there exists a universal threshold $\theta^*$ which, for \emph{any} data set, model, and adaptation procedure, always yields the best possible cumulative performance. We construct two data streams, both of total length $T$, that rely on the same pair of distinct concepts \(C_1\) and \(C_2\) but arrange them differently so as to expose a conflict for $\theta^*$. Reset model when drift detected and increment learning until model converge is chosen as adaption method in this proof.  

\textit{Stream $\mathcal{S}_1$.}
The first $t_d$ samples follow \(C_1\), and all subsequent samples (from index $t_d+1$ to $T$) follow \(C_2\). In such a sudden drift scenario, a relatively \emph{small} (i.e., sensitive) threshold is beneficial: triggering an immediate detection at $t_d+1$ sharply reduces mismatch time between model and data. By hypothesis, $\theta^*$ must be small enough to detect this drift right away so as to achieve optimal performance on $\mathcal{S}_1$.

\textit{Stream $\mathcal{S}_2$.}
In this second stream, $C_1$ again appears in the first $t_d$ samples, but then $C_2$ arises for \emph{exactly one} sample at $t_d+1$, after which the data revert to $C_1$ from index $t_d+2$ onward. Because $\theta^*$ was forced to be sufficiently sensitive to detect the appearance of $C_2$ in $\mathcal{S}_1$, the same threshold will also detect this one-sample drift in $\mathcal{S}_2$. Doing so, however, triggers a reset and adaptation to \(C_2\) that cannot complete before $C_1$ reappears at $t_d+2$. The model therefore faces another reset back to $C_1$, incurring two costly adaptation phases for a single $C_2$ sample. A larger threshold $\theta_2 > \theta^*$ could have ignored that transient drift, accepted one mismatch, and avoided these repeated resets, leading to higher overall accuracy on $\mathcal{S}_2$. Hence, $\theta^*$—which was optimal for $\mathcal{S}_1$—turns out to be suboptimal for $\mathcal{S}_2$, contradicting the claim that $\theta^*$ is universally optimal.

This conflict concludes the proof: no single threshold can perform optimally on all possible streams regardless of the choice of model and adaption method.
\end{proof}

\subsection{Proof of Theorem 3}

\begin{proof}
Divide the stream into substreams as \(D_1,\dots,D_N\) and each substreams contain only one concept drift. For each substream \(S_i\), define
\[
\theta_i^{*}
=
\arg\max_{\theta \in \Theta_{\mathrm{const}}}
\,A(\theta;\,D_i),
\quad
A_i^{*}
=
\max_{\theta \in \Theta_{\mathrm{const}}}
\,A(\theta;\,D_i)
=
A(\theta_i^{*};\,D_i).
\]
These \(\theta_i^{*}\) are the stationary thresholds that each maximize performance within the substream \(D_i\). Construct a dynamic strategy \(\{\theta_t\}\) by choosing \(\theta_t = \theta_i^{*}\) whenever \(t\) falls in substream \(D_i\). On substream \(D_i\), this choice attains performance \(A_i^{*}\). Summing over all \(i\),
\[
A(\{\theta_t\};\,D)
=
\sum_{i=1}^N
A(\theta_i^{*};\,D_i)
=
\sum_{i=1}^N
A_i^{*}.
\]

Take any stationary threshold \(\theta_{\mathrm{const}}\). On each substream \(D_i\), we have \(A(\theta_{\mathrm{const}};\,D_i)\le A_i^{*}\). Therefore,
\[
A(\theta_{\mathrm{const}};\,D)
=
\sum_{i=1}^N
A(\theta_{\mathrm{const}};\,D_i)
\le
\sum_{i=1}^N
A_i^{*}
=
A(\{\theta_t\};\,D).
\]
So the best dynamic strategy cannot be worse than the best single threshold.

If all \(\theta_i^{*}\) coincide, a single threshold matches the dynamic choice exactly, giving equal performance. Otherwise, there is at least one pair of substreams \(D_i\) and \(D_j\) for which \(\theta_i^{*}\neq \theta_j^{*}\). No single \(\theta_{\mathrm{const}}\) can match both \(A_i^{*}\) and \(A_j^{*}\). Hence that \(\theta_{\mathrm{const}}\) is strictly suboptimal in at least one substream, implying
\[
A(\theta_{\mathrm{const}};\,D)
<
\sum_{i=1}^N
A_i^{*}
=
A(\{\theta_t\};\,D),
\]
and the dynamic approach strictly outperforms any stationary threshold.
\end{proof}

\end{document}